\documentclass[letterpaper]{article} 
\usepackage[preprint]{aaai2027}  
\usepackage[hyphens]{url}  
\usepackage{graphicx} 
\usepackage{natbib}  
\usepackage{caption} 
\usepackage{amsmath}
\usepackage{amssymb}
\usepackage{amsfonts}
\usepackage{array}
\usepackage{booktabs}
\usepackage{enumitem}
\usepackage{multirow}
\usepackage{xspace}
\usepackage[table]{xcolor}
\definecolor{rowhl}{gray}{0.90}
\definecolor{grouphl}{gray}{0.85}
\renewcommand{\arraystretch}{0.95}

\newcommand{\method}{SV-WAM\xspace}

\title{SV-WAM: An Efficient Surround-View World-Action Model for End-to-End Autonomous Driving}
\author{
Jinyang Wang\textsuperscript{\rm 1,2},
Shiwei Li\textsuperscript{\rm 2}\corresponding\thanks{Project leader.},
Junjian Wang\textsuperscript{\rm 1},
Zhiqiang Deng\textsuperscript{\rm 2},
Jianbin Gao\textsuperscript{\rm 3},\\
Yihang Zhao\textsuperscript{\rm 1},
Liu Liu\textsuperscript{\rm 2},
Yongjia Zhao\textsuperscript{\rm 4},
Jinlong Chen\textsuperscript{\rm 5},
Huirui Xu\textsuperscript{\rm 1},\\
Yifeng Pan\textsuperscript{\rm 2},
Kangwei Liu\textsuperscript{\rm 2},
Fan Ren\textsuperscript{\rm 2},
Ji Tao\textsuperscript{\rm 2},
Minghao Yang\textsuperscript{\rm 1}\corresponding
}
\affiliations{
\textsuperscript{\rm 1}Institute of Automation, Chinese Academy of Sciences\\
\textsuperscript{\rm 2}Chongqing Changan Technology Co., Ltd.\\
\textsuperscript{\rm 3}Civil Aviation University of China\\
\textsuperscript{\rm 4}Beihang University\\
\textsuperscript{\rm 5}Guilin University of Electronic Technology\\[0.25em]
{\small\textbf{Project Page:} \pdfstartlink attr{/Border [0 0 0]} user{/Subtype /Link /A << /S /URI /URI (https://jinyangwang427.github.io/SV-WAM/) >>}\textcolor{blue}{\url{https://jinyangwang427.github.io/SV-WAM/}}\pdfendlink}
}

\begin{document}
\maketitle

\begin{abstract}
World models (WMs) have demonstrated strong potential for end-to-end autonomous driving by learning predictive representations of future scene dynamics. However, generating future videos during inference introduces substantial computational overhead, leading many recent driving WMs to adopt a single front camera as input for efficient deployment. This design restricts spatial coverage in safety-critical maneuvers such as lane changes, merges, and turns. To address this limitation, we propose \method, a surround-view world-action model (WAM) that preserves full six-camera observations while maintaining efficient inference. \method leverages future-video prediction as dense training supervision for action learning within a shared generative model, rather than as an inference-time output. At the core of this design is an action-centered causal mask that prevents action tokens from attending to future-video tokens during joint action-video denoising. Consequently, the video branch can be discarded at deployment, enabling efficient action-only planning. Furthermore, we introduce a differentiable drivable-area compliance regularizer that penalizes vehicle-footprint corners approaching or crossing drivable boundaries, improving planning safety and boundary awareness. Extensive experiments on the closed-loop NAVSIMv2 benchmark and the open-loop nuScenes benchmark demonstrate that \method achieves state-of-the-art planning performance with low inference latency and competitive zero-shot transfer capability.
\end{abstract}

\begin{figure}[t]
\centering
\includegraphics[width=\columnwidth]{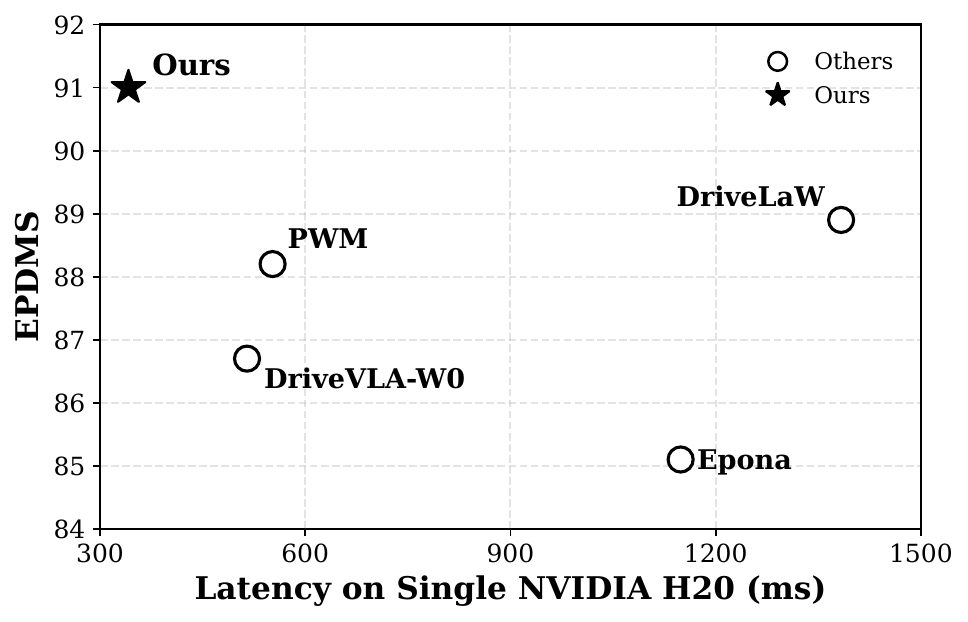}
\caption{Efficiency-performance overview on a single NVIDIA H20 GPU
(preprocessed-input-to-trajectory latency). \method reaches the best NAVSIMv2 EPDMS at the
lowest latency despite using six-camera input.}
\label{fig:teaser}
\end{figure}

\section{Introduction}

World models have recently emerged as a prominent approach for end-to-end autonomous driving~\cite{epona,pwm,eponav2,driveva,drivevla-w0,occworld,gaia1,drivedreamer,vista}. By forecasting the evolution of the driving scene, they provide planners with future-aware representations that extend beyond sparse action imitation~\cite{epona,pwm,eponav2,driveva,drivevla-w0}. This predictive supervision is particularly valuable for closed-loop planning, in which safe actions depend not only on the current observation but also on the constraints imposed by surrounding agents and road structure.

\begin{figure}[t]
\centering
\includegraphics[width=\columnwidth]{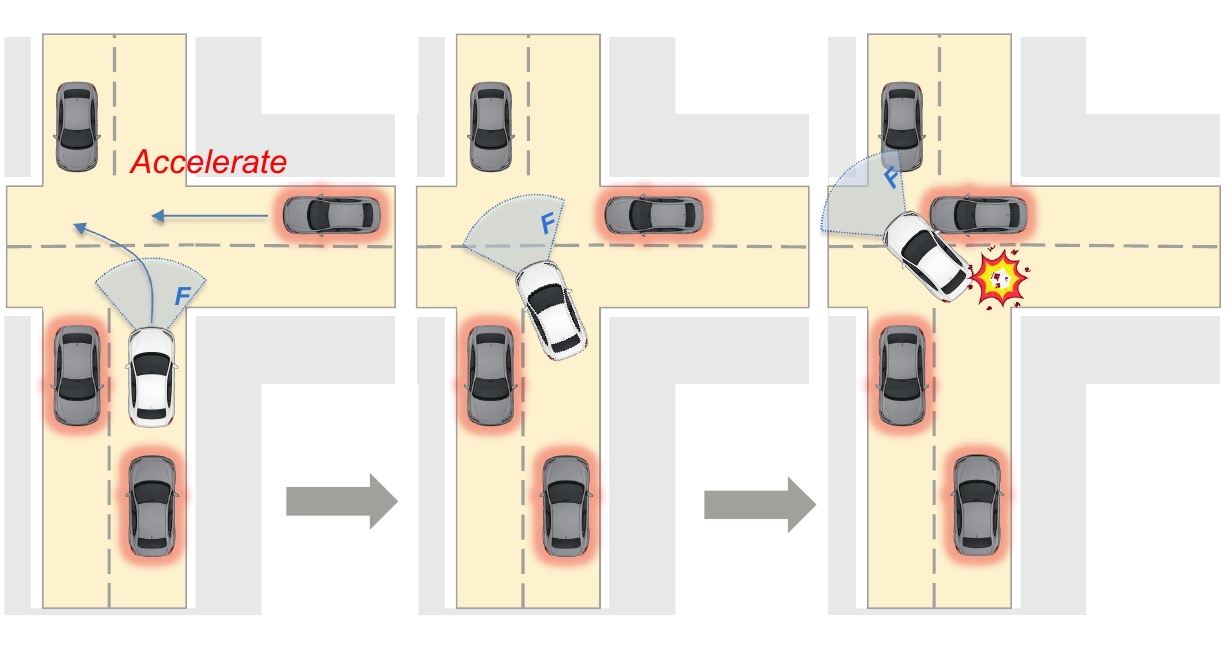}
\caption{Motivation for surround-view planning. For example, at unsignalized intersections, front-view-only planners may lack sufficient lateral context and make unsafe decisions in complex traffic.}
\label{fig:motivation}
\end{figure}

However, deploying world models for autonomous driving introduces a fundamental trade-off between reliability and computational efficiency. Reliable planning often requires surround-view observations, since side and rear context is critical for lane changes, merges, turns, and interactions with nearby vehicles. As illustrated in Figure~\ref{fig:motivation}, a front-view-only model may miss agents in lateral or rear blind zones, leading to unsafe decisions. Nevertheless, recent WM-based planners, such as Epona~\cite{epona}, PWM~\cite{pwm}, DriveVLA-W0~\cite{drivevla-w0}, and DriveLaW~\cite{drivelaw}, commonly adopt single-front-view inputs at inference to reduce the cost of future prediction and action generation. A straightforward extension to six-view future-video rollout would substantially increase the number of visual tokens and the inference latency. Therefore, the key challenge is how to retain surround-view evidence for reliable planning while avoiding expensive multi-view future-video generation at deployment.

To address this challenge, we propose \textbf{\method}, an efficient surround-view world-action model that uses future video as training-time supervision rather than an inference-time output. During training, \method jointly denoises future actions and six-view future-video latents within a shared diffusion-transformer backbone, allowing action prediction to benefit from future-scene dynamics through the shared model parameters. To avoid the inference-time computational cost of future-video prediction, we introduce an \textbf{action-centered causal mask} that prevents action tokens from attending to future-video tokens. This causal design allows the video branch to be removed at deployment, enabling efficient action-only inference while retaining full six-view spatial coverage. We further introduce a differentiable \textbf{drivable-area compliance regularizer} that penalizes vehicle footprints near or beyond drivable boundaries, thereby improving drivable-area compliance and planning safety without additional inference cost.

Experiments show that \method achieves strong planning performance with low latency. On NAVSIMv2, \method reaches \textbf{91.0 EPDMS} and runs in \textbf{342\,ms} on a single NVIDIA H20 GPU from preprocessed inputs to trajectory output, outperforming recent WM-based planners. On nuScenes, \method transfers \textbf{zero-shot} with \textbf{0.89\,m} average L2 error and \textbf{0.16\%} average collision rate.

Our contributions can be summarized as follows:
\begin{enumerate}[leftmargin=1.5em]
\item We propose \method, an efficient surround-view world-action model
with an action-centered causal mask that enables co-training of
future actions and surround-view future-video latents while retaining
action-only inference with full six-view conditioning.
\item We introduce a differentiable drivable-area regularizer that encourages predicted trajectories to remain within the drivable region, improving road compliance and safety without increasing inference cost.
\item Our method achieves state-of-the-art performance on NAVSIMv2 and demonstrates competitive zero-shot transfer on nuScenes with low latency.
\end{enumerate}

\section{Related Work}

\paragraph{End-to-End Autonomous Driving.}
End-to-end autonomous driving predicts future ego trajectories directly from raw sensor inputs~\cite{stp3}. BEV-based methods such as UniAD~\cite{uniad} and VAD~\cite{vad} integrate perception, prediction, and planning into a unified framework, while DiffusionDrive~\cite{diffusiondrive} improves multimodal trajectory generation with a truncated diffusion policy. Recent VLA-based planners further enhance driving with semantic reasoning; for example, ReCogDrive~\cite{recogdrive} connects scene-level understanding with continuous trajectory generation, and OpenDriveVLA~\cite{opendrivevla} grounds action prediction on 2D/3D visual representations and language commands~\cite{drivevlm,autovla,reflectdrive}. Despite their strong performance, many existing structured planners rely on multiple task-specific modules to construct explicit intermediate representations, such as object detections or occupancy maps, before trajectory generation. These modules increase architectural complexity. Moreover, they lack explicit modeling of future scene evolution.

\paragraph{Driving World Models for Planning.}
Driving world models have recently emerged for end-to-end planning~\cite{drivinggpt}. Epona~\cite{epona} learns an autoregressive diffusion world model that couples future scene prediction with trajectory planning, while PWM~\cite{pwm} integrates world modeling and trajectory planning through collaborative state--action prediction. DriveLaW~\cite{drivelaw} further unifies planning and video generation in a latent driving world, demonstrating the value of future visual prediction for downstream planning. However, these recent WM-based planners commonly use only a front view at inference, which limits spatial coverage in safety-critical maneuvers such as lane changes, merges, and turns. Recent latent world models extend this paradigm to three-front-view observations but rely on external perception priors. World4Drive~\cite{world4drive} uses pretrained spatial--semantic features for future-latent prediction and trajectory evaluation, while WorldRFT~\cite{worldrft} leverages vision--geometry priors for planning-aligned representation learning and reinforcement fine-tuning. Both retain pretrained geometric encoders in the inference path, introducing additional intermediate estimation stages. Directly extending such models to full surround-view inputs would further amplify the computational burden.

\paragraph{Efficient World-Action Models.}
Existing WAMs often follow imagine-then-execute or joint video--action modeling paradigms~\cite{unipi}, where future-video prediction remains in the inference loop and introduces substantial latency. Recent embodied WAMs therefore shift future prediction from test-time imagination to training-time supervision. Fast-WAM~\cite{fastwam} employs a
shared-attention Mixture-of-Transformers architecture that
couples a video DiT with an action expert DiT, retaining video
co-training while using the video branch only as a single-pass
world encoder at inference. GigaWorld-Policy~\cite{gigaworldpolicy} uses future-video generation as training-time supervision and retains only action prediction at test time. We bring this efficiency principle to surround-view driving, where preserving lateral and rear context must be balanced against the cost of multi-view future generation.

\section{Methodology}

\subsection{Overview}

\begin{figure*}[t]
  \centering
  \includegraphics[width=\textwidth]{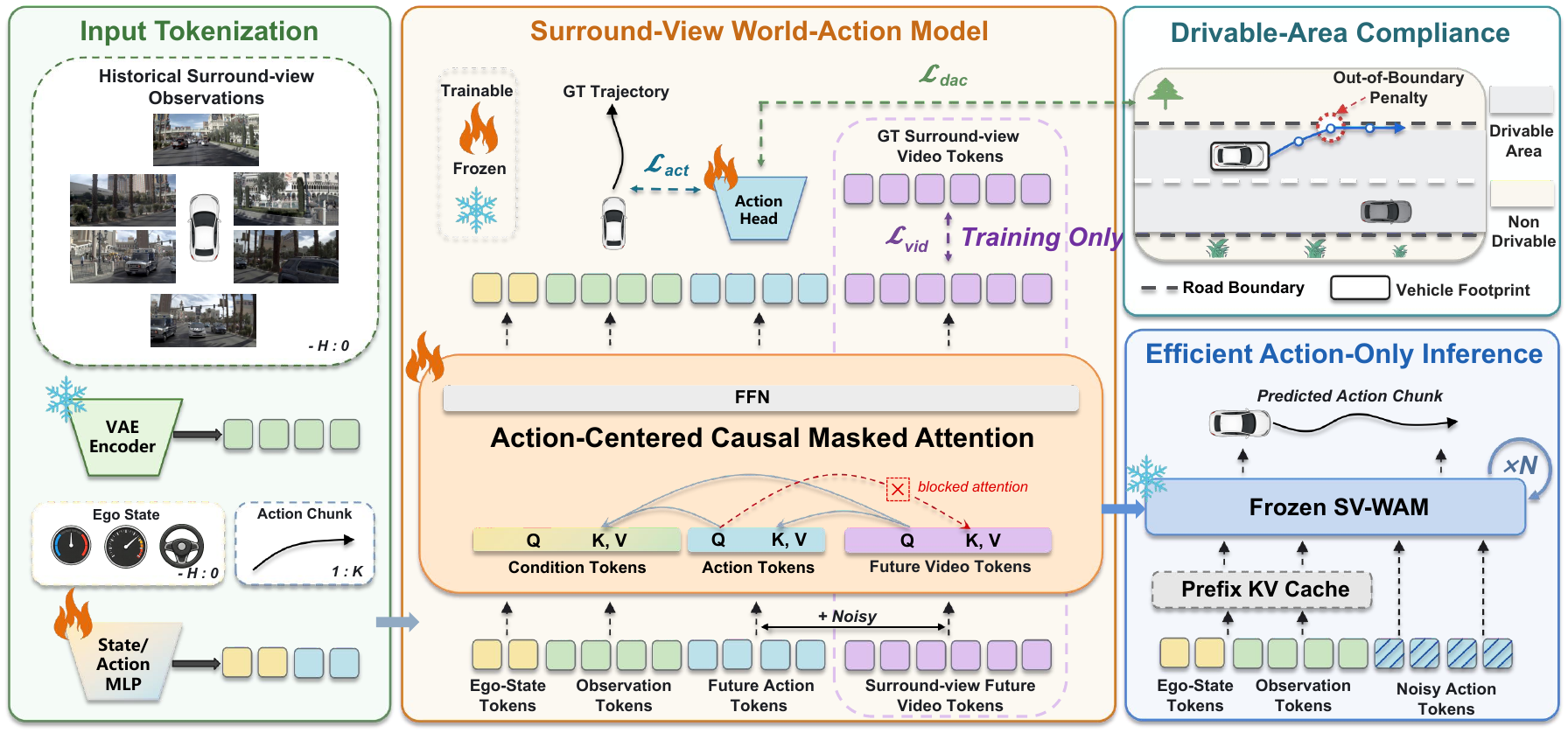}
\caption{Overview of \method.
During training, \method jointly denoises future actions and future-video
latents conditioned on six-view historical observations. The action-centered causal mask blocks action-token attention to
future-video tokens, while the
drivable-area compliance regularizer promotes road-compliant planning.
At inference, the future-video branch is removed, enabling efficient
action-only planning with full six-view context.}
  \label{fig:structure}
\end{figure*}

Figure~\ref{fig:structure} presents the overall framework of
\method, which consists of four components. First, input
tokenization converts historical surround-view observations,
ego states, and action sequences into transformer-compatible
tokens. Second, the surround-view world--action model jointly
denoises future actions and future-video latents using
an action-centered causal mask. Third, the drivable-area
compliance regularizer penalizes vehicle footprints that
approach or cross non-drivable boundaries during training.
Finally, at inference, \method removes the future-video branch
and reuses the cached condition prefix across denoising steps,
enabling efficient action-only planning.

\subsection{Input Tokenization}

This subsection describes how \method converts surround-view images, ego
states, actions, and future-video latents into transformer inputs.

\paragraph{Surround-View Video Encoding.}
Let $N_{\mathrm{cam}}=6$ denote the number of cameras. At each timestamp $i$,
the camera images are arranged in a fixed order and concatenated along the
width dimension:
\begin{equation}
  X_i=\operatorname{Concat}_{w}
  (I_i^1,I_i^2,\ldots,I_i^{N_{\mathrm{cam}}}),
\end{equation}
where $I_i^v$ denotes the image from the $v$-th camera and $X_i$ is the
resulting surround-view frame.

We encode the full training video sequence $X_{-H+1:F}$ using the frozen 3D
causal VAE encoder of Wan2.2-TI2V-5B~\cite{wan22}. The resulting latents are
split along the temporal dimension:
\begin{equation}
  [Z^{\mathrm{ref}},Z^{\mathrm{fut}}]
  =
  E_{\mathrm{vae}}(X_{-H+1:F}).
\end{equation}
Here, $H$ and $F$ denote the numbers of historical and future
frames, respectively; $Z^{\mathrm{ref}}$ represents the clean historical prefix used for
visual conditioning, while $Z^{\mathrm{fut}}$ contains the future-video
latents used only during training. Both are tokenized using a shared video
tokenizer:
\begin{equation}
  [\mathcal{V}^{\mathrm{ref}},\mathcal{V}^{\mathrm{fut}}]
  =
  \operatorname{Tok}_v
  ([Z^{\mathrm{ref}},Z^{\mathrm{fut}}]).
\end{equation}
In implementation, $\operatorname{Tok}_v$ partitions the latents into
spatiotemporal patches and embeds them into a flattened transformer-token
sequence.

\paragraph{State and Action Encoding.}
Following the NAVSIM input convention~\cite{navsim}, we denote the
ego-state history as $S=\{s_i\}_{i=-H+1}^{0}$, where
$s_i\in\mathbb{R}^{8}$. An MLP maps the state history into state tokens:
\begin{equation}
  \mathcal{S}=\phi_s(S).
\end{equation}

The future action sequence is written as
$A=\{a_k=(\Delta x_k,\Delta y_k,\Delta\psi_k)\}_{k=1}^{K}$, where each action
represents a stepwise ego-relative motion increment. The sequence is projected
into action tokens using an MLP tokenizer:
\begin{equation}
  \mathcal{A}=\phi_a(A).
\end{equation}

\subsection{Surround-View World-Action Model}
This subsection introduces the action-centered causal mask mechanism
for efficient inference and the joint flow-matching training used in \method.
\paragraph{Action-Centered Causal Mask.}

Recent video-action driving models jointly generate future visual forecasts and action sequences in a shared generative process~\cite{driveva}. Inspired by embodied world-action modeling~\cite{gigaworldpolicy}, we design an action-centered causal mask for efficient planning. The mask blocks the direct attention dependency of action tokens on future-video tokens, which are expensive to denoise and are not instantiated in the action-only inference path.

The mask is applied over the packed token sequence during self-attention. During flow training, we sample a shared base time
$u\sim\mathcal{U}(0,1)$ and apply a shifted flow schedule
to obtain $t$ for both action and future-video denoising. Given Gaussian noises $\epsilon_a$ and $\epsilon_z$, we construct
\begin{equation}
  A_t=(1-t)A_0+t\epsilon_a,
  \qquad
  Z_t^{\mathrm{fut}}=(1-t)Z_0^{\mathrm{fut}}+t\epsilon_z,
\end{equation}
where $A_0$ and $Z_0^{\mathrm{fut}}$ denote the clean future action sequence and
clean future video latents. Using the tokenizers defined in the previous subsection, $A_t$ and
$Z_t^{\mathrm{fut}}$ are converted into noisy action tokens $\mathcal{A}_t$
and noisy future-video tokens $\mathcal{V}^{\mathrm{fut}}_t$. The resulting self-attention sequence is
\begin{equation}
  \mathcal{Y}_t
  =
  [\mathcal{S},\mathcal{V}^{\mathrm{ref}},
  \mathcal{A}_t,\mathcal{V}^{\mathrm{fut}}_t].
\end{equation}

\begin{figure}[t]
  \centering
  \includegraphics[width=\columnwidth]{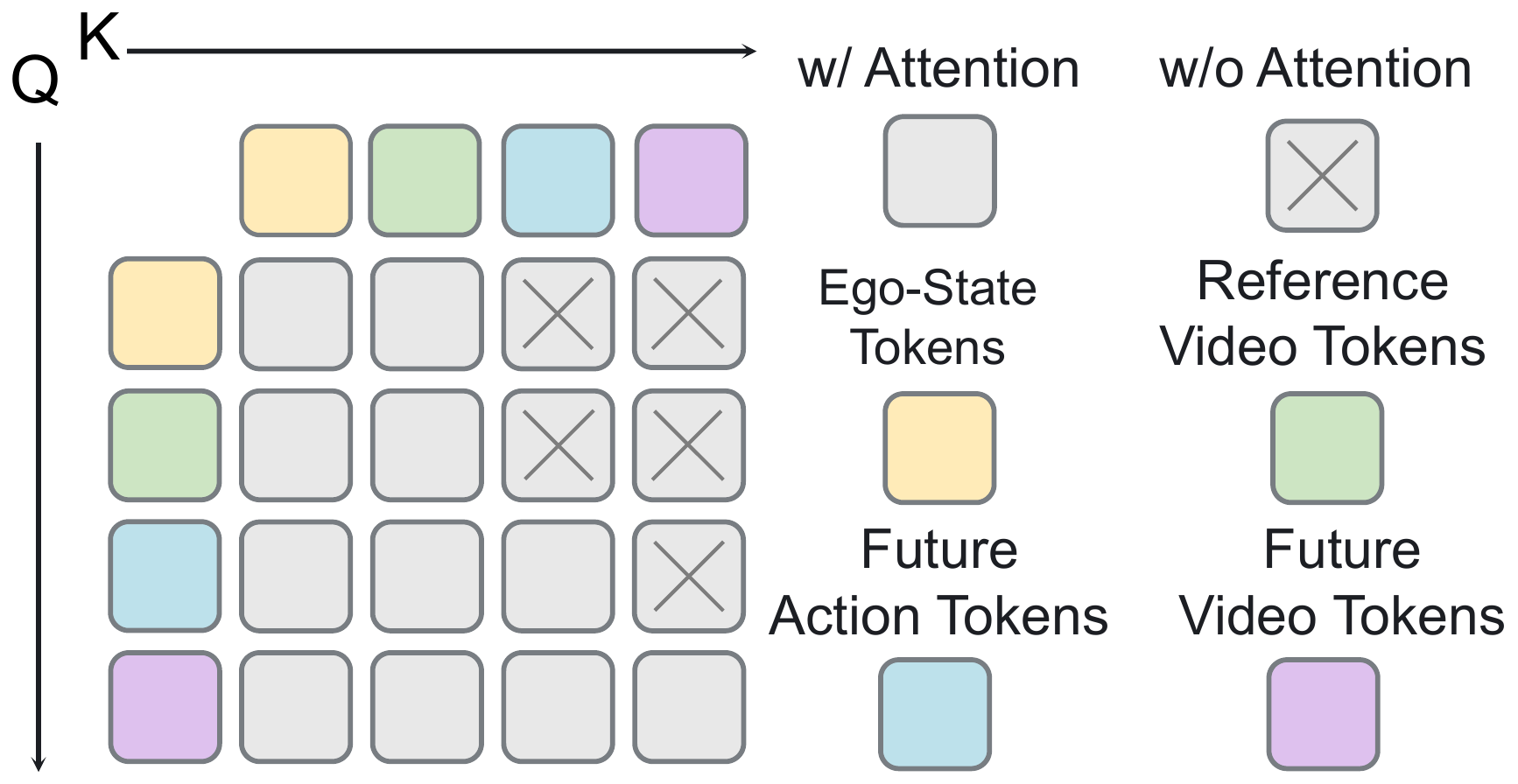}
\caption{Action-centered causal mask.
Action tokens attend to ego-state and reference-video tokens but are blocked from attending to future-video tokens, whereas future-video tokens can attend to action tokens. This asymmetric dependency supports future-video co-training without making action prediction dependent on future-video tokens at inference.}
  \label{fig:mask}
\end{figure}

As shown in Figure~\ref{fig:mask}, state tokens and reference video tokens form the clean condition prefix and attend only to themselves. Action tokens attend to the condition prefix and action tokens, while future video tokens attend to the condition prefix, action tokens, and future video tokens. 

Let $\Omega=\{Z^{\mathrm{ref}},S\}$ denote the condition consisting of reference video latents and ego-state history. The mask induces an action-centered factorization:
\begin{equation}
  p_\theta(A_0,Z_0^{\mathrm{fut}}\mid\Omega)
  =
  p_\theta(A_0\mid\Omega)
  p_\theta(Z_0^{\mathrm{fut}}\mid\Omega,A_0).
\end{equation}
Thus, future-video denoising provides future-dynamics
supervision through the shared model parameters, while action
tokens do not attend to future-video tokens in the forward
pass. This property enables action-only inference.

\paragraph{\method Training.}
We train \method with flow matching~\cite{flow-matching,rectified-flow} on both future actions and future video latents. Given the noisy variables defined above, the target velocity fields for the two branches are obtained by differentiating the interpolations $A_t$ and $Z_t^{\mathrm{fut}}$ with respect to $t$:
\begin{equation}
  v_a^\star=\epsilon_a-A_0,
  \qquad
  v_z^\star=\epsilon_z-Z_0^{\mathrm{fut}}.
\end{equation}

The masked transformer takes the self-attention sequence $\mathcal{Y}_t$ and the flow time $t$ as input:
\begin{equation}
  (\hat v_a,\hat v_z)
  =
  f_\theta(\mathcal{Y}_t,t),
\end{equation}
where $f_\theta$ denotes the masked velocity predictor with learnable parameters $\theta$, and $\hat v_a$ and $\hat v_z$ are the predicted action- and video-velocity fields. The action velocity is decoded from the action token positions using an MLP action head, while the video velocity is projected from the future video token positions and unpatchified back to the VAE latent space.

The flow matching loss is defined as
\begin{equation}
\begin{aligned}
  \mathcal{L}_{\mathrm{fm}}
  ={}&
  \lambda_{\mathrm{act}}
  \left\|\hat v_a-v_a^\star\right\|_2^2 \\
  &+
  \lambda_{\mathrm{vid}}
  \left\|\hat v_z-v_z^\star\right\|_2^2.
\end{aligned}
\end{equation}
The action branch directly optimizes trajectory generation, while the future video branch encourages the model to capture future scene dynamics. Because the causal mask prevents action-token queries from attending to future-video keys and values, future-video supervision introduces no inference-time dependency into the action path.

\subsection{Drivable Area Compliance Regularization}

Flow matching learns future ego actions from trajectory supervision, but it does not explicitly constrain the simulated vehicle footprint with respect to drivable boundaries. Motivated by the drivable area compliance metric in NAVSIM~\cite{navsim}, we propose a drivable area compliance regularizer that softly penalizes predicted trajectories whose footprint approaches or leaves the drivable region. Given the predicted action velocity $\hat v_a$, we first recover the clean action prediction from the noisy action input:
\begin{equation}
  \hat A_0 = A_t - t\hat v_a.
\end{equation}
Given a binary drivable area map $b$, we construct a signed distance field $\Phi$ in the ego-centric coordinate system:
\begin{equation}
  \Phi(u)=D_{\mathrm{non}}(u)-D_{\mathrm{drive}}(u),
\end{equation}
where $D_{\mathrm{non}}(u)$ and $D_{\mathrm{drive}}(u)$ denote the Euclidean distances from location $u$ to the nearest non-drivable and drivable cells, respectively. Under this convention, $\Phi(u)>0$ indicates that $u$ lies inside the drivable area, while $\Phi(u)<0$ indicates an off-road location.

After denormalization, we follow the official NAVSIM evaluation pipeline~\cite{navsim} and use its LQR-based tracker to roll out $\hat A_0$ into simulated ego poses $Q=\{q_i\}_{i=1}^{N_q}$, where $q_i=(x_i,y_i,\psi_i)$.
 For each simulated pose, we compute the four corner points $c_{i,j}$ of the vehicle footprint and sample their signed distances from the signed distance field:
\begin{equation}
  d_{i,j}=\Phi(c_{i,j}),
  \qquad
  j=1,\ldots,4.
\end{equation}
Since the corner points are continuous coordinates, $\Phi(c_{i,j})$ is evaluated by bilinear interpolation on the signed distance field grid.

We apply a smooth margin penalty to each sampled distance:
\begin{equation}
  \ell(d_{i,j})
  =
  \beta
  \log\left(
  1+
  \exp\left(
  \frac{m-d_{i,j}}{\beta}
  \right)
  \right),
\end{equation}
where $m$ is the safety margin and $\beta$ controls the smoothness. The final drivable area compliance loss is computed by log-mean-exp aggregation over all footprint penalties:
\begin{equation}
  \mathcal{L}_{\mathrm{dac}}
  =
  \rho
  \log\left(
  \frac{1}{4N_q}
  \sum_{i=1}^{N_q}
  \sum_{j=1}^{4}
  \exp\left(
  \frac{\ell(d_{i,j})}{\rho}
  \right)
  \right),
\end{equation}
where $\rho$ is the aggregation temperature. This aggregation acts as a smooth maximum and assigns larger weights to high-risk footprint violations than simple averaging. The overall training objective is
\begin{equation}
  \mathcal{L}
  =
  \mathcal{L}_{\mathrm{fm}}
  +
  \lambda_{\mathrm{dac}}\mathcal{L}_{\mathrm{dac}}.
\end{equation}

\subsection{Efficient Action-Only Inference}

At inference time, we keep the same reference video and state conditioning path but omit noisy future video tokens from the transformer sequence. The inference sequence becomes
\begin{equation}
  \mathcal{Y}^{\mathrm{act}}_t
  =
  [\mathcal{S},\mathcal{V}^{\mathrm{ref}},\mathcal{A}_t].
\end{equation}
The same trained model is evaluated under the corresponding action-only mask:
\begin{equation}
  \hat v_a
  =
  f_\theta
  (\mathcal{Y}^{\mathrm{act}}_t,t),
\end{equation}
where $f_\theta$ denotes the trained action-velocity predictor. Starting from Gaussian action noise, the action scheduler iteratively updates the action sequence using the predicted velocity field. Since noisy future video tokens are not instantiated and the video output branch is skipped, \method avoids future video denoising and decoding during deployment while retaining the representation benefit of training-time video co-denoising. To further improve efficiency, we cache and reuse the conditioning prefix across denoising steps.

\section{Experiments}
\label{sec:experiments}

\subsection{Experimental Setup}

\paragraph{Benchmarks and Metrics.}
We evaluate closed-loop planning on NAVSIMv2~\cite{navsim,pseudo-simulation}, which reports the
Extended Predictive Driver Model Score (EPDMS). EPDMS aggregates no at-fault
collision (NC), drivable-area compliance (DAC), driving-direction compliance (DDC),
traffic-light compliance (TLC), ego progress (EP), time-to-collision (TTC), lane
keeping (LK), history comfort (HC), and extended comfort (EC), where the compliance
terms act as multiplicative safety penalties. To test cross-dataset generalization,
we further evaluate on nuScenes~\cite{nuscenes} without target-domain fine-tuning.
For nuScenes, we report L2 trajectory error and collision rate at 1s, 2s, 3s, and
their averages. NAVSIMv1 and NAVSIMv2 navhard results are reported in the
Appendix.

\paragraph{Implementation Details.}
Our planner uses \(H=4\) historical frames from six cameras: front-left, front,
front-right, rear-right, rear, and rear-left. Each view is resized to
\(448\times224\), and the six views are concatenated along the width dimension
before being encoded by the frozen 3D causal VAE encoder from Wan2.2-TI2V-5B.
Both historical observations and future-video targets are sampled at
\(2\,\mathrm{Hz}\) (a \(0.5\,\mathrm{s}\) temporal interval).
We adopt Wan2.2-5B as the DiT backbone~\cite{dit}. The model predicts 12 future action tokens
and is trained with 12-frame six-view future video latent supervision. We train on
NAVSIM trainval for 10k iterations with AdamW, batch size 80, learning rate
\(10^{-4}\), and weight decay \(10^{-2}\) on 16 H800 GPUs. We then fine-tune for 1k
iterations with an effective batch size of 640 (gradient accumulation of 8 over a
per-step batch of 80) and a learning rate decayed from \(10^{-5}\) to \(10^{-6}\);
each stage takes roughly one day on the 16 GPUs. All reported main results use
two-step action-only inference. Future video tokens provide training-time supervision
only. At inference, we remove the video branch and cache the observation prefix as
key/value states.

\subsection{Main Results}

\paragraph{NAVSIMv2.}

As demonstrated in Table~\ref{tab:navsim_v2_navtest_detail}, \method achieves
the best performance under the closed-loop NAVSIMv2 metric suite, outperforming
recent end-to-end driving planners on the navtest split. We follow the evaluation setting of
EponaV2~\cite{eponav2}. The results for DriveVLA-W0 and DriveLaW are reproduced using
their official code and checkpoints, while the remaining baseline results are taken
from the corresponding papers or official leaderboard. \method achieves the best
EPDMS of 91.0. These gains
should not be attributed to surround-view input alone. The six-view observations
provide complementary side and rear context, while future-video co-training and the
drivable-area regularizer further improve future-aware representation learning and
road-boundary compliance, respectively, as analyzed in
Table~\ref{tab:component_ablation}.

\begin{table*}[t]
\centering
\begingroup
\small
\setlength{\tabcolsep}{0.5pt}
\renewcommand{\arraystretch}{0.90}
\begin{tabular*}{\textwidth}{@{}l@{\extracolsep{\fill}}cc*{9}{c}>{\columncolor{grouphl}}c@{}}
\toprule
Method & Ref. & Input & NC \(\uparrow\) & DAC \(\uparrow\) & DDC \(\uparrow\) & TLC \(\uparrow\) &
EP \(\uparrow\) & TTC \(\uparrow\) & LK \(\uparrow\) & HC \(\uparrow\) & EC \(\uparrow\) & EPDMS \(\uparrow\) \\
\midrule
ReCogDrive~\cite{recogdrive} & ICLR'26 & C\(\times\)1 & 98.3 & 95.2 & \underline{99.5} & \underline{99.8} & 87.1 & 97.5 & 96.6 & \underline{98.3} & 86.5 & 83.6 \\
Epona~\(\dagger\)~\cite{epona} & ICCV'25 & C\(\times\)1 & 97.1 & 95.7 & 99.3 & 99.7 & \textbf{88.6} & 96.3 & 97.0 & 98.0 & 67.8 & 85.1 \\
DriveVLA-W0~\(\dagger\)~\cite{drivevla-w0} & ICLR'26 & C\(\times\)1 & 98.3 & 95.0 & 99.4 & \textbf{99.9} & 86.6 & 97.8 & \underline{97.7} & \textbf{98.4} & 82.4 & 86.7 \\
PWM~\(\dagger\)~\cite{pwm} & NeurIPS'25 & C\(\times\)1 & \underline{98.8} & 95.9 & 99.4 & \textbf{99.9} & 86.4 & \underline{98.4} & 97.6 & \underline{98.3} & 85.3 & 88.2 \\
DriveLaW~\(\dagger\)~\cite{drivelaw} & CVPR'26 & C\(\times\)1 & \underline{98.8} & 96.9 & \textbf{99.6} & \underline{99.8} & 87.4 & 98.3 & \underline{97.7} & \textbf{98.4} & 80.2 & 88.9 \\
EponaV2~\(\dagger\)~\cite{eponav2} & arXiv'26 & C\(\times\)1 & 98.5 & 97.4 & \underline{99.5} & \textbf{99.9} & 87.9 & 98.1 & \underline{97.7} & 98.2 & 77.4 & 88.9 \\
\textbf{Ours}~\(\dagger\) & -- & C\(\times\)1 & 98.4 & 98.4 & 99.1 & \underline{99.8} & 85.5 & 97.8 & 95.9 & \textbf{98.4} & 86.5 & 89.3 \\
\midrule
AutoDrive-P3~\cite{autodrive-p3} & ICLR'26 & C\(\times\)3 & \textbf{99.1} & 97.4 & 99.2 & \underline{99.8} & \underline{88.0} & \textbf{98.7} & 96.3 & \underline{98.3} & 85.5 & 89.9 \\
WoTE~\(\dagger\)~\cite{wote} & ICCV'25 & C\(\times\)3\&L & 98.5 & 96.8 & 98.8 & \underline{99.8} & 86.1 & 97.9 & 95.5 & \underline{98.3} & 82.9 & 87.7 \\
DiffusionDrive~\cite{diffusiondrive} & CVPR'25 & C\(\times\)3\&L & 98.2 & 96.2 & \underline{99.5} & \underline{99.8} & 87.4 & 97.3 & 96.9 & \textbf{98.4} & \underline{87.7} & 88.2 \\
\textbf{Ours}~\(\dagger\) & -- & C\(\times\)3 & 98.7 & \textbf{99.1} & 99.4 & \underline{99.8} & 86.1 & 98.1 & 96.5 & \textbf{98.4} & 86.6 & \underline{90.5} \\
\midrule
\textbf{Ours}~\(\dagger\) & -- & C\(\times\)6 & 98.6 & \underline{98.8} & \textbf{99.6} & \textbf{99.9} & 86.8 & 98.3 & \textbf{97.8} & \textbf{98.4} & \textbf{88.4} & \textbf{91.0} \\
\bottomrule

\end{tabular*}
\endgroup
\caption{Detailed closed-loop planning comparison on the NAVSIMv2 navtest split with human penalty enabled.
Metrics are reported as percentages. C denotes camera input and L denotes LiDAR input;
for example, C\(\times\)1 is front-camera-only input, C\(\times\)3 denotes three front-facing camera
input, and C\(\times\)6 denotes six-view surround-camera input. Bold and underlined values indicate the best and second-best results overall, respectively. \(\dagger\) denotes WM-based methods.}
\label{tab:navsim_v2_navtest_detail}
\end{table*}

\paragraph{nuScenes Zero-Shot Transfer.}

As shown in Table~\ref{tab:nuscenes_merged}, we further evaluate zero-shot transfer to
nuScenes. The upper block lists representative methods fine-tuned on nuScenes, whereas the lower block
compares world-model planners without nuScenes fine-tuning. We directly evaluate the
officially released checkpoints of DriveVLA-W0 and PWM without further fine-tuning on
NAVSIM or nuScenes. Under this setting, \method achieves the lowest average L2 error
of 0.89\,m and collision rate of 0.16\% among the zero-shot methods, while remaining
competitive with several in-domain fine-tuned planners. This result provides encouraging evidence of cross-dataset
transfer without target-domain supervision. Additional
zero-shot qualitative results on in-house driving data are provided in the
Appendix.

\begin{table*}[t]
\centering
\begingroup
\small
\setlength{\tabcolsep}{0pt}
\renewcommand{\arraystretch}{0.90}
\begin{tabular*}{\textwidth}{@{}l@{\extracolsep{\fill}}cccccc>{\columncolor{grouphl}}cccc>{\columncolor{grouphl}}c@{}}
\toprule
\multirow{2}{*}{Method} & \multirow{2}{*}{nuS.\ FT} & \multirow{2}{*}{Ref} &
\multirow{2}{*}{Input} &
\multicolumn{4}{c}{L2 (m) \(\downarrow\)} &
\multicolumn{4}{c}{Collision Rate (\%) \(\downarrow\)} \\
\cmidrule(lr){5-8}\cmidrule(lr){9-12}
 & & & & 1s & 2s & 3s & Avg. & 1s & 2s & 3s & Avg. \\
\midrule
UniAD~\cite{uniad} & \checkmark & CVPR'23 & C\(\times\)6 & 0.48 & 0.96 & 1.65 & 1.03 & 0.05 & 0.17 & 0.71 & 0.31 \\
VAD-Base~\cite{vad} & \checkmark & ICCV'23 & C\(\times\)6 & 0.54 & 1.15 & 1.98 & 1.22 & \underline{0.04} & 0.39 & 1.17 & 0.53 \\
GenAD~\cite{genad} & \checkmark & ECCV'24 & C\(\times\)6 & \underline{0.36} & \underline{0.83} & \textbf{1.55} & \underline{0.91} & 0.06 & 0.23 & 1.00 & 0.43 \\
Doe-1~\cite{doe1} & \checkmark & arXiv'24 & C\(\times\)1 & 0.50 & 1.18 & 2.11 & 1.26 & \underline{0.04} & 0.37 & 1.19 & 0.53 \\
Epona~\(\dagger\)~\cite{epona} & \checkmark & ICCV'25 & C\(\times\)1 & 0.61 & 1.17 & 1.98 & 1.25 & \textbf{0.01} & 0.22 & 0.85 & 0.36 \\
DriveLaW~\(\dagger\)~\cite{drivelaw} & \checkmark & CVPR'26 & C\(\times\)1 & 0.44 & 1.10 & 1.91 & 1.15 & 0.15 & \textbf{0.10} & \underline{0.48} & \underline{0.24} \\
\midrule
DriveVLA-W0~\(\dagger\)~\cite{drivevla-w0} & \(\times\) & ICLR'26 & C\(\times\)1 & 0.43 & 1.26 & 2.60 & 1.43 & 0.22 & 0.66 & 1.42 & 0.77 \\
PWM~\(\dagger\)~\cite{pwm} & \(\times\) & NeurIPS'25 & C\(\times\)1 & 2.06 & 3.91 & 6.00 & 3.99 & 0.12 & 0.15 & 0.86 & 0.36 \\
\textbf{Ours}~\(\dagger\) & \(\times\) & -- & C\(\times\)6 & \textbf{0.31} & \textbf{0.80} & \underline{1.57} & \textbf{0.89} & 0.07 & \underline{0.13} & \textbf{0.27} & \textbf{0.16} \\
\bottomrule
\end{tabular*}
\endgroup
\caption{End-to-end motion planning performance on nuScenes. ``nuS.\ FT'' marks
whether the method is fine-tuned on nuScenes. DriveVLA-W0 and PWM are evaluated zero-shot on nuScenes using their officially released checkpoints, without additional fine-tuning. Results for additional methods are provided in the Appendix. Bold and underlined values indicate the best and second-best results, respectively. \(\dagger\) denotes WM-based methods.}
\label{tab:nuscenes_merged}
\end{table*}
\begin{table}[t]
\centering
\begingroup
\small
\renewcommand{\arraystretch}{0.90}
\begin{tabular*}{\columnwidth}{@{}c@{\extracolsep{\fill}}cccc@{}}
\toprule
Video Sup. & DAC-reg & FT & DAC \(\uparrow\) & EPDMS \(\uparrow\) \\
\midrule
\(\times\) & \(\times\) & \(\times\) & 93.4 & 83.1 \\
\checkmark & \(\times\) & \(\times\) & 95.6 & 87.7 \\
\checkmark & \checkmark & \(\times\) & 98.6 & 90.1 \\
\checkmark & \checkmark & \checkmark & \textbf{98.8} & \textbf{91.0} \\
\bottomrule
\end{tabular*}
\endgroup
\caption{Component ablation on NAVSIMv2 navtest. \emph{Video Sup.} denotes
future-video supervision co-training, \emph{DAC-reg} denotes the drivable-area
regularizer, and \emph{FT} denotes final fine-tuning.}
\label{tab:component_ablation}
\end{table}

\subsection{Ablation Studies}
\paragraph{Training Components.}
Table~\ref{tab:component_ablation} ablates the three ingredients of the final model. All variants share the same C$\times$6 architecture and evaluation
setup, differing only in the enabled training components. Future-video co-training improves EPDMS from 83.1 to 87.7, showing that grounding
actions in future visual dynamics helps even though the video branch is removed at
deployment. Adding the DAC regularization raises EPDMS to 90.1 and DAC from 95.6 to 98.6, indicating fewer drivable-area violations. Final fine-tuning yields 91.0 EPDMS.

\paragraph{Action-Centered Causal Mask and Inference Steps.}
Table~\ref{tab:mask_ablation} jointly studies the action-centered causal mask and the
number of inference steps. For a controlled comparison, the masked and unmasked variants are trained with
identical model architectures, training configurations, and hyperparameters,
differing only in the attention mask. Without the mask, action tokens can read noisy future-video
tokens during joint generation. Increasing the steps from 2 to 10 improves EPDMS from
87.3 to 88.2, consistent with more complete future-video denoising reducing the noise
that propagates into action prediction. With the causal mask, action tokens no longer
depend on future-video tokens; two steps attain the best EPDMS of 91.0, while 5 and 10
steps give 90.9 and 90.8. Additional refinement is therefore unnecessary for the
decoupled action path, and its small decline likely reflects minor iterative
perturbations of an already converged action trajectory rather than useful correction
from the uncertain future-video branch. Moreover, we ablate the future-video branch at inference under the causal
mask and observe unchanged planning performance with or without it.

\begin{figure*}[t]
\centering
\includegraphics[width=\textwidth]{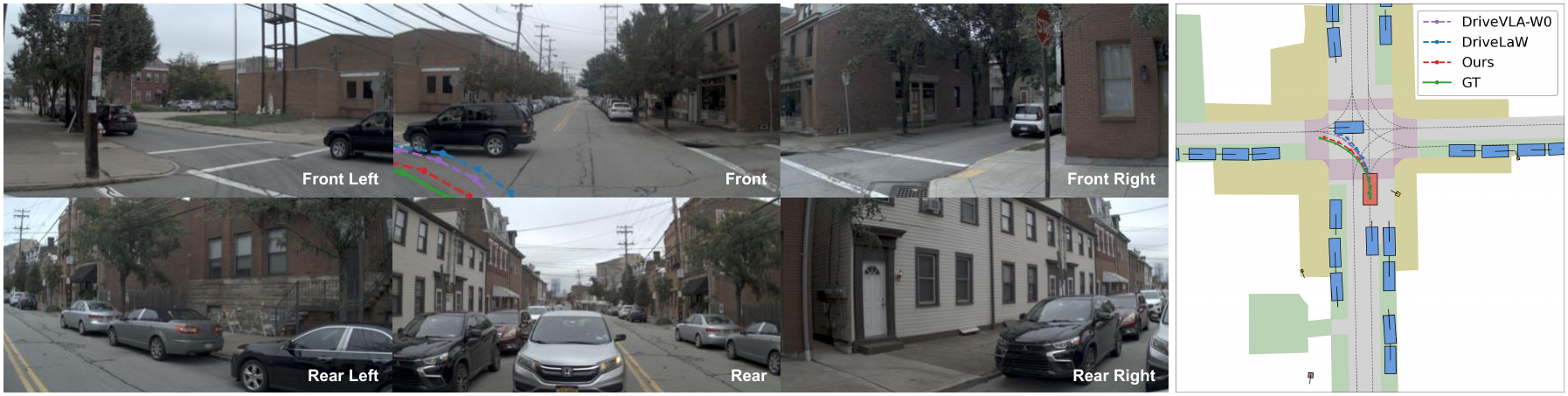}
\caption{Qualitative comparison in a challenging turning scenario
from NAVSIMv2. Left: current six-camera observations. Right: the corresponding BEV
visualization. \method predicts a trajectory that more closely follows the
ground truth and better aligns with the road geometry than DriveLaW and
DriveVLA-W0.}
\label{fig:comparison_visual}
\end{figure*}

\begin{table*}[t]
\centering
\begingroup
\small
\setlength{\tabcolsep}{0pt}
\renewcommand{\arraystretch}{0.90}
\begin{tabular*}{\textwidth}{@{}c@{\extracolsep{\fill}}cc*{10}{c}@{}}
\toprule
Causal Mask & Steps & Input & NC \(\uparrow\) & DAC \(\uparrow\) & DDC \(\uparrow\) &
TLC \(\uparrow\) & EP \(\uparrow\) & TTC \(\uparrow\) & LK \(\uparrow\) &
HC \(\uparrow\) & EC \(\uparrow\) & EPDMS \(\uparrow\) \\
\midrule
\(\times\) & 2  & C\(\times\)6 & 97.4 & 96.6 & 99.0 & 99.7 & \textbf{87.8} & 96.8 & 97.0 & 98.3 & 86.0 & 87.3 \\
\(\times\) & 5  & C\(\times\)6 & 97.6 & 96.8 & 99.2 & 99.7 & 87.7 & 96.9 & 97.5 & 98.3 & 87.3 & 87.9 \\
\(\times\) & 10 & C\(\times\)6 & 97.7 & 96.9 & 99.3 & 99.7 & 87.6 & 97.0 & 97.6 & 98.3 & 87.3 & 88.2 \\
\midrule
\checkmark & 2  & C\(\times\)6 & \textbf{98.6} & \textbf{98.8} & 99.6 & \textbf{99.9} & 86.8 & 98.3 & 97.8 & \textbf{98.4} & \textbf{88.4} & \textbf{91.0} \\
\checkmark & 5  & C\(\times\)6 & \textbf{98.6} & 98.6 & 99.6 & \textbf{99.9} & 86.8 & \textbf{98.4} & \textbf{97.9} & \textbf{98.4} & \textbf{88.4} & 90.9 \\
\checkmark & 10 & C\(\times\)6 & \textbf{98.6} & 98.5 & \textbf{99.7} & 99.8 & 86.9 & 98.2 & \textbf{97.9} & \textbf{98.4} & \textbf{88.4} & 90.8 \\
\bottomrule
\end{tabular*}
\endgroup
\caption{Ablation on the causal mask and inference steps on NAVSIMv2 navtest. Bold values indicate the best results.}
\label{tab:mask_ablation}
\end{table*}

\paragraph{Surround-View Coverage.}
The controlled variants in Table~\ref{tab:navsim_v2_navtest_detail} share the
same architecture and training schedule. Increasing camera coverage consistently improves EPDMS, with the full six-view setting
achieving the best final score. The improvements in DDC, LK, EC, and EPDMS indicate
that wider visual coverage benefits overall planning quality and several safety-related
submetrics. This trend is consistent with the
motivation for surround-view inputs, as side and rear observations provide
complementary context that is unavailable from a single front camera.

\subsection{Efficiency Analysis}

Figure~\ref{fig:teaser} summarizes the efficiency--performance trade-off against
recent world-model-based planners. Table~\ref{tab:speed} compares their online
trajectory-output latency on a single H20 GPU. The timing starts from preprocessed inputs to final trajectory. For the compared methods, we use
official code, released checkpoints, and default inference configurations. All methods
use batch size 1 and bf16 where supported; 30 samples remain after 5 warm-up samples.
Despite using six cameras, \method achieves the lowest latency at
\(341.6\pm2.1\)\,ms. Removing the action-centered causal mask requires the future-video branch
to remain active at inference, increasing the trajectory-output latency to \(848.0 \pm 1.8\) ms, with optional video decoding excluded in both cases. The full measurement protocol and optional video-rendering costs are provided in the Appendix.

\begin{table}[t]
\centering
\begingroup
\small
\setlength{\tabcolsep}{0pt}
\renewcommand{\arraystretch}{0.90}
\begin{tabular*}{\columnwidth}{@{}l@{\extracolsep{\fill}}cccc@{}}
\toprule
Method & Vis. Enc. & Core Gen. & Other & \shortstack{Total (ms) \(\downarrow\)\\mean\(\pm\)std} \\
\midrule
Epona & 725.4 & 415.2 & 8.2 & 1148.8\(\pm\)4.5 \\
DriveLaW & 459.7 & 663.7 & 259.9 & 1383.3\(\pm\)19.7 \\
PWM & 210.2 & 336.0 & 6.1 & 552.3\(\pm\)5.4 \\
DriveVLA-W0 & 76.6 & 369.0 & 69.3 & 514.9\(\pm\)3.8 \\
Ours (w/o mask) & 438.4 & 392.5 & 17.1 & 848.0\(\pm\)1.8 \\
\textbf{Ours} & 174.7 & 151.4 & 15.5 & \textbf{341.6\(\pm\)2.1} \\
\bottomrule
\end{tabular*}
\endgroup
\caption{Online trajectory-output latency comparison among WM-based planning methods on a single NVIDIA H20 GPU. Vis. Enc., Core Gen., and Other denote visual tokenization,
iterative backbone generation, and remaining pipeline overhead,
respectively. Component-wise values are means in milliseconds, while total
latencies are reported as mean \(\pm\) standard deviation. Optional video rendering and future-video VAE decoding are excluded.}
\label{tab:speed}
\end{table}
\subsection{Qualitative Analysis}

Figure~\ref{fig:comparison_visual} compares \method with DriveLaW and
DriveVLA-W0 in a challenging turning scenario. Benefiting from the richer
spatial context provided by six-view observations and the drivable-area
compliance regularizer, \method predicts a trajectory that more closely follows
the ground-truth path and remains within the drivable corridor,
demonstrating stronger planning robustness in complex maneuvers. Additional
qualitative comparisons across diverse driving scenarios are provided in the
Appendix.

\section{Conclusion}
We present \method, an efficient surround-view world-action model for end-to-end autonomous driving. The key insight is that six-camera context should be retained for reliable planning, whereas future video prediction can be used as training-time supervision and removed from inference. \method follows this insight with an action-centered causal mask that enables joint denoising of future actions and future-video latents while blocking the direct attention dependency of action tokens on future-video tokens. Future-video supervision still updates the shared model parameters, while the video branch can be removed at inference, yielding an efficient action-only planner with full surround-view coverage. We further introduce a differentiable drivable area compliance regularizer to improve road compliance. Experiments on NAVSIMv2 and nuScenes validate the effectiveness and efficiency of the proposed design.

\bibliography{references}
\clearpage
\appendix
\providecommand{\refm}[1]{\textcolor{black!55}{#1}}
\providecommand{\secbest}[1]{\underline{#1}}

\section{Implementation Details}
\label{app:impl}

\paragraph{Architecture.}
\method builds on the Wan2.2-5B diffusion-transformer (DiT) video backbone and the
3D causal VAE encoder from Wan2.2-TI2V-5B. The four-frame six-view history is encoded by the VAE after the six views
at each timestamp are concatenated along the width dimension into a single mosaic, and
the resulting latents are converted into video tokens by a 3D convolutional patch embedding followed by flattening. Ego states and
actions are projected into the transformer hidden space by lightweight MLP tokenizers,
and a linear head maps the action token outputs back to the flow-velocity space. The
transformer keeps the Wan2.2-5B DiT configuration unchanged---30 blocks and
\(\approx\!5\)B parameters, with the original hidden width and head count---and the only
additions are the small action and ego-state MLP tokenizers, whose parameter count is
negligible.

\paragraph{Inputs and Outputs.}
Each sample uses \(H=4\) history frames from the six cameras (front-left, front,
front-right, rear-right, rear, rear-left), each resized to \(448\times224\). The
ego state at each history step is
\begin{equation}
  s_t=[c_t^1,c_t^2,c_t^3,c_t^4,v_t^x,v_t^y,a_t^x,a_t^y]\in\mathbb{R}^{8},
\end{equation}
where \(c_t\) is the four-dimensional one-hot driving command. The command entries
remain in their native \(\{0,1\}\) space. We min--max normalize the continuous
components to \([-1,1]\) using statistics computed on the official NAVSIM training
split. In the order \((v^x,v^y,a^x,a^y)\), the extrema are
\begin{equation}
  \begin{aligned}
    s_{\min}&=(-2.0,-0.5,-5.1,-5.4),\\
    s_{\max}&=(20.4,0.5,3.6,5.6).
  \end{aligned}
\end{equation}
Each continuous component is transformed as
\(\tilde{s}=2(s-s_{\min})/(s_{\max}-s_{\min})-1\) and clipped to \([-1,1]\).

Unlike the ego state, the action targets are not normalized with dataset statistics.
Inspired by the coordinate-wise bounded normalization of Epona~\cite{epona}, we use
fixed ranges adapted to our stepwise \(0.5\,\mathrm{s}\) ego-relative action
representation. Each action is \(a_k=(\Delta x_k,\Delta y_k,\Delta\psi_k)\), with
nominal bounds
\begin{equation}
  a_{\min}=(0,-0.5,-8^{\circ}),\qquad
  a_{\max}=(8,0.5,8^{\circ}),
\end{equation}
where the translational components are in meters. We apply
\(\tilde{a}=2(a-a_{\min})/(a_{\max}-a_{\min})-1\), without clipping values outside
the nominal range. The model predicts \(K=12\) future action tokens, which are
denormalized and accumulated in \(\mathrm{SE}(2)\); the first eight tokens form the
standard \(4\,\mathrm{s}\) NAVSIM trajectory. During training, the model additionally
denoises 12 future six-view video latents.

\paragraph{Training.}
We train on the official NAVSIM trainval split for 10k iterations with AdamW (batch size 80, learning
rate \(10^{-4}\), weight decay \(10^{-2}\)) on 16 H800 GPUs, then fine-tune for 1k
iterations with an effective batch size of 640 (gradient accumulation of 8 over a
per-step batch of 80) and a learning rate decayed from \(10^{-5}\) to \(10^{-6}\). Each
stage takes roughly one day (\(\approx\!24\)h) on the 16 H800 GPUs. We use rectified-flow training with Euler sampling; the full \(\approx\!5\)B model is trainable.
The flow matching loss uses
\(\lambda_{\mathrm{act}}:\lambda_{\mathrm{vid}}=1:1\). The implementation samples a uniform base time $u$ and applies
the shifted flow schedule
\begin{equation}
  t=\frac{\alpha u}{1+(\alpha-1)u},\qquad u\sim\mathcal{U}(0,1),
\end{equation}
where the flow-shift coefficient is set to
$\alpha=5.0$ for both action and future-video flow matching. The drivable area regularizer uses
\(\lambda_{\mathrm{dac}}=0.01\), safety margin \(m=0.2\), softplus temperature
\(\beta=0.2\), and log-mean-exp aggregation temperature \(\rho=0.1\). The signed
distance field is constructed on a local grid
\(x\in[-20,80]\,\mathrm{m}\), \(y\in[-40,40]\,\mathrm{m}\) at \(0.5\,\mathrm{m}\)
resolution.

\paragraph{Drivable Area Regularization Details.}
The predicted \(2\,\mathrm{Hz}\) action sequence specifies a geometric reference
rather than the trajectory attained by a tracking controller. We therefore roll it
out with a linear quadratic regulator (LQR) and a kinematic bicycle
model~\cite{kinematic-bicycle}, following NAVSIM's official tracker~\cite{navsim}. LQR is a
finite-horizon feedback controller that balances reference-tracking accuracy against
control effort. In its standard form, it selects a control sequence according to
\begin{equation}
  u_{0:H-1}^{\star}
  =\arg\min_{u_{0:H-1}}
  \sum_{h=0}^{H-1}
  \left(e_h^{\top}Qe_h+u_h^{\top}Ru_h\right),
\end{equation}
where \(e_h\) denotes the tracking-error state and \(u_h\) denotes the control. In
our rollout, the predicted poses at \(0.5\,\mathrm{s}\) intervals are first
interpolated to the \(0.1\,\mathrm{s}\) simulation grid. The longitudinal controller
tracks the reference velocity through acceleration commands, while the lateral LQR
uses lateral displacement, heading, and steering-angle errors to determine the
steering-rate command. First-order actuator dynamics are applied before propagating
the 11-dimensional vehicle state with the kinematic bicycle model. We simulate
\(N_q=60\) poses with an LQR horizon of 10 steps.

NAVSIM's official tracker, simulator, and drivable-area routines operate on NumPy
arrays and therefore cannot preserve gradients from the compliance objective to the
action generator. We retain the state definition, controller structure, vehicle
geometry, and drivable-map semantics of the official implementation, but reimplement
the rollout and footprint queries as batched PyTorch tensor operations. Specifically,
we obtain the binary drivable region using the official
PDMDrivableMap, convert it into an ego-centric signed-distance tensor, and
evaluate the four vehicle-corner points at every simulated pose by differentiable
bilinear interpolation. Out-of-bound samples are assigned negative distances. The
softplus margin penalty and log-mean-exp aggregation described in the main paper are
then evaluated entirely in PyTorch, allowing \(\mathcal{L}_{\mathrm{dac}}\) to
backpropagate through the LQR rollout to the predicted actions.

\paragraph{Inference.}
At deployment, the future-video branch is removed. The observation prefix is encoded once and its key/value states are cached, and the 12 action tokens are denoised in two flow steps. When future-video generation is needed for analysis, noisy future-video latent tokens can be appended to the input and denoised by the same model, after which the predicted latents are decoded into future video. Because the action-centered causal mask prevents action tokens from attending to future-video tokens, this optional video branch does not affect the action path. We empirically verify that, under the same number of inference steps, the masked model produces identical trajectories up to numerical precision with and without future-video tokens. This observation is consistent with the causal masking design and supports action-only deployment. The main-paper latency numbers are measured on a single NVIDIA H20 with batch size 1 and bf16 precision; additional details are provided in the inference-efficiency section.

\section{Dataset and Metric Details}
\label{app:datasets}

\subsection{NAVSIM}

\paragraph{Overview.}
The NAVSIM benchmark family is built on OpenScene, a \(2\,\mathrm{Hz}\) subset of the nuPlan driving logs with approximately 120 hours of real-world driving data. OpenScene provides the standard trainval, test, and mini splits. The trainval split contains \(14\,\mathrm{GB}\) of log files and more than \(2000\,\mathrm{GB}\) of sensor data, the test split contains \(1\,\mathrm{GB}\) of log files and \(217\,\mathrm{GB}\) of sensor data, and the mini split contains \(1\,\mathrm{GB}\) of log files and \(151\,\mathrm{GB}\) of sensor data. NAVSIM further defines filtered planning splits over these data to emphasize non-trivial driving scenes. Specifically, navtest is derived from test and is used for standardized NAVSIMv1 evaluation. In our experiments, we train \method on the NAVSIM trainval data and report the main results on the NAVSIMv2~\cite{navsim,pseudo-simulation} navtest split.

NAVSIMv2 extends the original NAVSIM benchmark with pseudo-simulation~\cite{pseudo-simulation}. Instead of evaluating only the logged observation at the initial state, NAVSIMv2 uses pre-generated synthetic observations to approximate shifted future states that a planner may encounter after deviating from the expert trajectory. This design allows the benchmark to evaluate error recovery and causal-confusion sensitivity while avoiding expensive sequential sensor rendering. Besides the main navtest evaluation, we also report results on the harder navhard two-stage split, which contains real Stage 1 observations and synthetic Stage 2 observations and applies a two-stage evaluation protocol with human penalty. Finally, we include NAVSIMv1 navtest results under the original PDMS metric to connect our results with earlier NAVSIM-based work.

\paragraph{Data Processing.}
Each raw camera view in NAVSIM has a resolution of \(1920 \times 1080\) pixels. We first
resize each selected camera view to \(455 \times 256\) pixels and then center-crop it to
\(448 \times 224\) pixels. At each timestamp, the six processed camera views are arranged
in the fixed order described above and concatenated along the width dimension into
a single surround-view mosaic. We use four history frames sampled at \(2\,\mathrm{Hz}\),
corresponding to a temporal interval of \(0.5\,\mathrm{s}\).

\paragraph{NAVSIMv2 Metric (EPDMS).}
The Extended Predictive Driver Model Score (EPDMS) aggregates multiplicative
safety penalties and weighted quality terms,
\begin{equation}
\begin{aligned}
\mathrm{EPDMS}
&=
\underbrace{
\mathrm{NC}\cdot\mathrm{DAC}\cdot\mathrm{DDC}\cdot\mathrm{TLC}
}_{\text{multiplicative safety}}
\\[-1mm]
&\quad \cdot
\frac{
5(\mathrm{EP}{+}\mathrm{TTC})+
2(\mathrm{LK}{+}\mathrm{HC}{+}\mathrm{EC})
}{16}.
\end{aligned}
\end{equation}
Here the four safety terms act as multiplicative gates: NC (no at-fault collision)
penalizes collisions for which the ego is responsible, DAC (drivable-area compliance)
checks that the rolled-out vehicle footprint stays inside the drivable area, DDC
(driving-direction compliance) penalizes wrong-way or illegal-direction motion, and TLC
(traffic-light compliance) penalizes running red lights. The remaining terms are quality
scores combined as a weighted average: EP (ego progress) measures route progress
relative to a privileged reference, TTC (time-to-collision) rewards keeping a safe
temporal margin to other agents, LK (lane keeping) measures staying within the lane, and
HC (history comfort) and EC (extended comfort) bound acceleration and jerk for ride
comfort. Because NC, DAC, DDC, and TLC enter multiplicatively, a single violation (for
example, leaving the drivable area) zeroes the entire scenario score; this is exactly the
failure mode our drivable-area regularizer targets. The navtest split contains 12{,}146
scenarios and the navhard split 450 scenarios (official sizes).

\paragraph{NAVSIMv2 navhard.}
The navhard split (Table~\ref{tab:navsim_v2_navhard_detail}) is a harder NAVSIMv2 setting
that targets interactive, safety-critical situations. It uses a two-stage evaluation: the
planner is first scored on the recorded scene and then re-scored under more challenging
conditions, and an additional human-behavior penalty is applied. Because the second stage
stresses reactive behavior, navhard scores are markedly lower than navtest for all
methods, so the split mainly measures robustness in difficult scenes.

\paragraph{NAVSIMv1 metric (PDMS).}
The NAVSIMv1 results in Table~\ref{tab:navsim_v1_detail} use the Predictive Driver Model
Score (PDMS), the predecessor of EPDMS,
\begin{equation}
\mathrm{PDMS}=
\underbrace{\mathrm{NC}\cdot\mathrm{DAC}}_{\text{multiplicative safety}}
\cdot
\frac{5\,\mathrm{EP}+5\,\mathrm{TTC}+2\,\mathrm{C}}{12},
\end{equation}
where NC and DAC are multiplicative safety terms and EP, TTC, and C (comfort) form a
weighted average. Compared with EPDMS, PDMS omits the additional v2 terms (DDC, TLC, LK)
and does not split comfort into history and extended comfort.

\subsection{nuScenes}

\paragraph{Overview.}
nuScenes~\cite{nuscenes} contains 1000 driving scenes of roughly \(20\,\mathrm{s}\) each,
collected in Boston and Singapore with a synchronized six-camera surround rig and
keyframe annotations at \(2\,\mathrm{Hz}\). It spans diverse urban traffic, lighting, and
weather conditions, which makes it a strong testbed for cross-dataset generalization. We
use it purely for evaluation and never train or fine-tune on it.

\paragraph{Zero-Shot Protocol.}
For zero-shot evaluation, the compared world-model planners, including DriveVLA-W0,
PWM, and \method, are trained only on NAVSIM and evaluated directly on nuScenes
without target-domain fine-tuning. We evaluate on the nuScenes validation split using the standard open-loop planning
protocol. For \method, we use the four most recent keyframes at \(2\,\mathrm{Hz}\) from all
six cameras. Each native RGB image has a resolution of \(1600 \times 900\) pixels. We
bilinearly resize each image to \(455 \times 256\) pixels and then center-crop it to
\(448 \times 224\) pixels, matching the deterministic preprocessing used during NAVSIM
training. The six processed views are concatenated along the width dimension in
the same fixed order. The provided planning command is mapped to our four-way
command encoding, while ego velocity and acceleration are normalized using the
NAVSIM training statistics. We neither compute target-domain normalization
statistics nor perform any parameter adaptation on nuScenes.

\paragraph{Metrics.}
Following the standard nuScenes open-loop planning evaluation used in UniAD~\cite{uniad} and subsequent planning works~\cite{vad}, we report L2 displacement error and collision rate at 1\,s, 2\,s, and 3\,s. The L2 displacement error measures the Euclidean distance between the predicted ego position and the ground-truth ego position at each future horizon, in meters. The collision rate measures the percentage of predicted trajectories that collide with annotated scene agents within each evaluation horizon. We also report the average L2 error and average collision rate over the three horizons.

\section{Additional Quantitative Results}
\label{app:additional_navsim}

We report three comparisons omitted from the main paper for space: the NAVSIMv1 benchmark
(Table~\ref{tab:navsim_v1_detail}), the harder NAVSIMv2 navhard split
(Table~\ref{tab:navsim_v2_navhard_detail}), and the complete nuScenes comparison with
older in-domain baselines (Table~\ref{tab:nuscenes_full}).

\paragraph{NAVSIMv1.}
On the NAVSIMv1 PDMS benchmark (Table~\ref{tab:navsim_v1_detail}), \method reaches
90.2 PDMS, outperforming DriveLaW (89.1), PWM (88.1), and Epona (86.2), while achieving
98.6 DAC. These results further support the effectiveness of the proposed planner and
drivable-area regularization.

\paragraph{NAVSIMv2 navhard.}
On the harder navhard split (Table~\ref{tab:navsim_v2_navhard_detail}), which combines a
two-stage evaluation with a human-behavior penalty, \method reaches 36.1 EPDMS, tying
EponaV2 and ranking behind RAP (39.6) among the compared methods. It also achieves 94.9
first-stage DAC and strong extended-comfort scores (EC, 84.4 and 80.1 across the two
stages), indicating robust behavior under the more challenging reactive conditions.

\paragraph{nuScenes (full).}
Table~\ref{tab:nuscenes_full} extends the main nuScenes comparison with older in-domain
baselines (ST-P3, OccNet, OccWorld, VAD-Tiny). Although \method is evaluated zero-shot,
its average L2 (0.89\,m) and collision rate (0.16\%) are competitive with, and on average
better than, several nuScenes-fine-tuned baselines, indicating that the surround-view
world-action representation transfers across datasets. Because the open-loop nuScenes
L2/collision metric is known to be sensitive to ego-status shortcuts, we treat this as a
cross-dataset sanity check rather than a headline result.

\begin{table*}[t]
\centering
\begingroup
\small
\renewcommand{\arraystretch}{0.90}
\setlength{\tabcolsep}{0pt}
\begin{tabular*}{\textwidth}{@{}l@{\extracolsep{\fill}}*{5}{c}>{\columncolor{grouphl}}c@{}}
\toprule
Method & NC \(\uparrow\) & DAC \(\uparrow\) & EP \(\uparrow\) & TTC \(\uparrow\) & C \(\uparrow\) & PDMS \(\uparrow\) \\
\midrule
TransFuser~\cite{transfuser} & 97.7 & 92.8 & 79.2 & 92.8 & 100.0 & 84.0 \\
DiffusionDrive~\cite{diffusiondrive} & 98.2 & 96.2 & 82.2 & 94.7 & 100.0 & 88.1 \\
VGGDrive~\cite{vggdrive} & 98.6 & 96.3 & 82.9 & 95.6 & 100.0 & 88.8 \\
ResWorld~\cite{resworld} & 98.9 & 96.5 & 83.1 & 95.6 & 100.0 & 89.0 \\
MeanFuser~\cite{meanfuser} & 98.6 & 97.0 & 82.8 & 95.0 & 100.0 & 89.0 \\
AutoVLA~\cite{autovla} & 98.4 & 95.6 & 81.9 & 98.0 & 99.9 & 89.1 \\
DriveVLA-W0 (Anchor)~\cite{drivevla-w0} & 98.7 & 99.1 & 83.3 & 95.3 & 99.3 & 90.2 \\
AutoDrive-P3~\cite{autodrive-p3} & 99.1 & 97.4 & 84.8 & 96.5 & 100.0 & 90.6 \\
ReCogDrive~\cite{recogdrive} & 97.9 & 97.3 & 87.3 & 94.9 & 100.0 & 90.8 \\
SafeDrive~\cite{safedrive} & 99.5 & 99.0 & 84.3 & 97.2 & 100.0 & 91.6 \\
DrivingGPT~\cite{drivinggpt} & 98.9 & 90.7 & 79.7 & 94.9 & 95.6 & 82.4 \\
World4Drive~\cite{world4drive} & 97.4 & 94.3 & 79.9 & 92.8 & 100.0 & 85.1 \\
WorldRFT~\cite{worldrft} & 97.8 & 96.8 & 81.7 & 94.0 & 100.0 & 87.8 \\
Epona~\cite{epona} & 97.9 & 95.1 & 80.4 & 93.8 & 99.9 & 86.2 \\
DriveVLA-W0 (Flow)~\cite{drivevla-w0} & 98.4 & 95.3 & 80.9 & 95.2 & 100.0 & 87.2 \\
PWM~\cite{pwm} & 98.6 & 95.9 & 81.8 & 95.4 & 100.0 & 88.1 \\
DriveVLA-W0 (Query)~\cite{drivevla-w0} & 98.7 & 96.2 & 82.2 & 95.5 & 100.0 & 88.4 \\
DriveLaW~\cite{drivelaw} & 99.0 & 97.1 & 81.3 & 96.7 & 100.0 & 89.1 \\
EponaV2~\cite{eponav2} & 98.6 & 97.9 & 84.8 & 95.7 & 100.0 & 90.4 \\
Ours & 98.6 & 98.6 & 83.9 & 95.1 & 100.0 & 90.2 \\
\bottomrule
\end{tabular*}
\endgroup
\caption{Detailed comparison on the NAVSIMv1 benchmark~\cite{navsim}.
NC, DAC, EP, TTC, and C are the standard Predictive Driver Model Score (PDMS)
components.}
\label{tab:navsim_v1_detail}
\end{table*}

\begin{table*}[t]
\centering
\begingroup
\small
\setlength{\tabcolsep}{0pt}
\renewcommand{\arraystretch}{0.90}
\begin{tabular*}{\textwidth}{@{}l@{\extracolsep{\fill}}c*{10}{c}@{}}
\toprule
Method & Stage & NC \(\uparrow\) & DAC \(\uparrow\) & DDC \(\uparrow\) & TLC \(\uparrow\) &
EP \(\uparrow\) & TTC \(\uparrow\) & LK \(\uparrow\) & HC \(\uparrow\) & EC \(\uparrow\) & EPDMS \(\uparrow\) \\
\midrule
\multirow{2}{*}{LTF~\cite{transfuser}}
& 1 & 96.2 & 79.6 & 99.1 & 99.6 & 84.1 & 95.1 & 94.2 & 97.6 & 79.1 & \multirow{2}{*}{25.1} \\
& 2 & 77.8 & 70.2 & 84.3 & 98.1 & 85.1 & 75.7 & 45.4 & 95.7 & 76.0 & \\
\cmidrule[0.22pt](lr){1-12}

\multirow{2}{*}{LTFv6~\cite{lead}}
& 1 & 96.6 & 86.7 & 99.2 & 99.6 & 84.5 & 95.1 & 94.4 & 97.8 & 76.4 & \multirow{2}{*}{31.9} \\
& 2 & 79.9 & 75.6 & 86.3 & 97.9 & 89.6 & 76.1 & 50.1 & 95.2 & 66.7 & \\
\cmidrule[0.22pt](lr){1-12}

\multirow{2}{*}{RAP~\cite{rap}}
& 1 & 97.1 & 94.4 & 98.8 & 99.8 & 83.9 & 96.9 & 94.7 & 96.4 & 66.2 & \multirow{2}{*}{39.6} \\
& 2 & 83.2 & 83.9 & 87.4 & 98.0 & 86.9 & 80.4 & 52.3 & 95.2 & 52.4 & \\
\multirow{2}{*}{DriveVLA-W0~\cite{drivevla-w0}}
& 1 & 96.8 & 83.3 & 99.0 & 99.6 & 84.6 & 95.3 & 96.4 & 97.6 & 78.2 & \multirow{2}{*}{24.4} \\
& 2 & 76.8 & 64.3 & 79.9 & 98.3 & 89.2 & 75.0 & 46.8 & 95.8 & 53.1 & \\
\cmidrule[0.22pt](lr){1-12}
\multirow{2}{*}{DriveLaW~\cite{drivelaw}}
& 1 & 97.3 & 89.1 & 99.2 & 99.6 & 84.3 & 97.1 & 96.2 & 97.8 & 67.6 & \multirow{2}{*}{30.6} \\
& 2 & 82.5 & 67.6 & 83.5 & 98.1 & 84.8 & 78.5 & 45.8 & 96.4 & 57.3 & \\
\cmidrule[0.22pt](lr){1-12}
\multirow{2}{*}{EponaV2~\cite{eponav2}}
& 1 & 97.3 & 90.7 & 99.4 & 100.0 & 83.3 & 97.3 & 97.3 & 97.6 & 60.9 & \multirow{2}{*}{36.1} \\
& 2 & 83.6 & 78.0 & 88.0 & 98.9 & 86.0 & 80.3 & 50.1 & 96.1 & 52.0 & \\
\cmidrule[0.22pt](lr){1-12}
\multirow{2}{*}{Ours}
& 1 & 96.6 & 94.9 & 98.8 & 99.8 & 82.5 & 96.9 & 97.1 & 97.8 & 84.4 & \multirow{2}{*}{36.1} \\
& 2 & 82.9 & 75.8 & 84.3 & 98.0 & 82.1 & 79.1 & 48.2 & 96.4 & 80.1 & \\
\bottomrule
\end{tabular*}
\endgroup
\caption{Detailed comparison on the NAVSIMv2 navhard split with human penalty enabled.
Each method is evaluated in two stages; EPDMS is the final split score.}
\label{tab:navsim_v2_navhard_detail}
\end{table*}

\begin{table*}[t]
\centering
\begingroup
\small
\setlength{\tabcolsep}{0pt}
\renewcommand{\arraystretch}{0.90}

\begin{tabular*}{\textwidth}{
@{}l@{\extracolsep{\fill}}
ccc
ccc>{\columncolor{grouphl}}c
ccc>{\columncolor{grouphl}}c@{}
}
\toprule
\multirow{2}{*}{Method}
& \multirow{2}{*}{nuS.\ FT}
& \multirow{2}{*}{Ref}
& \multirow{2}{*}{Input}
& \multicolumn{4}{c}{L2 (m) \(\downarrow\)}
& \multicolumn{4}{c}{Collision Rate (\%) \(\downarrow\)} \\
\cmidrule(lr){5-8}
\cmidrule(lr){9-12}
& & & &
1s & 2s & 3s & Avg. &
1s & 2s & 3s & Avg. \\
\midrule

ST-P3~\cite{stp3}
& \checkmark & ECCV'22 & C\(\times\)6
& 1.33 & 2.11 & 2.90 & 2.11
& 0.23 & 0.62 & 1.27 & 0.71 \\

UniAD~\cite{uniad}
& \checkmark & CVPR'23 & C\(\times\)6
& 0.48 & 0.96 & 1.65 & 1.03
& 0.05 & 0.17 & 0.71 & 0.31 \\

OccNet~\cite{occnet}
& \checkmark & ICCV'23 & C\(\times\)6
& 1.29 & 2.13 & 2.99 & 2.14
& 0.21 & 0.59 & 1.37 & 0.72 \\

OccWorld~\cite{occworld}
& \checkmark & ECCV'24 & C\(\times\)6
& 0.52 & 1.27 & 2.41 & 1.40
& 0.12 & 0.40 & 2.08 & 0.87 \\

VAD-Tiny~\cite{vad}
& \checkmark & ICCV'23 & C\(\times\)6
& 0.60 & 1.23 & 2.06 & 1.30
& 0.31 & 0.53 & 1.33 & 0.72 \\

VAD-Base~\cite{vad}
& \checkmark & ICCV'23 & C\(\times\)6
& 0.54 & 1.15 & 1.98 & 1.22
& \underline{0.04} & 0.39 & 1.17 & 0.53 \\

GenAD~\cite{genad}
& \checkmark & ECCV'24 & C\(\times\)6
& \underline{0.36} & \underline{0.83} & \textbf{1.55}
& \underline{0.91}
& 0.06 & 0.23 & 1.00 & 0.43 \\

Doe-1~\cite{doe1}
& \checkmark & arXiv'24 & C\(\times\)1
& 0.50 & 1.18 & 2.11 & 1.26
& \underline{0.04} & 0.37 & 1.19 & 0.53 \\

Epona~\cite{epona}
& \checkmark & ICCV'25 & C\(\times\)1
& 0.61 & 1.17 & 1.98 & 1.25
& \textbf{0.01} & 0.22 & 0.85 & 0.36 \\

DriveLaW~\cite{drivelaw}
& \checkmark & CVPR'26 & C\(\times\)1
& 0.44 & 1.10 & 1.91 & 1.15
& 0.15 & \textbf{0.10} & \underline{0.48}
& \underline{0.24} \\

\midrule

DriveVLA-W0~\cite{drivevla-w0}
& \(\times\) & ICLR'26 & C\(\times\)1
& 0.43 & 1.26 & 2.60 & 1.43
& 0.22 & 0.66 & 1.42 & 0.77 \\

PWM~\cite{pwm}
& \(\times\) & NeurIPS'25 & C\(\times\)1
& 2.06 & 3.91 & 6.00 & 3.99
& 0.12 & 0.15 & 0.86 & 0.36 \\

Ours
& \(\times\) & -- & C\(\times\)6
& \textbf{0.31} & \textbf{0.80} & \underline{1.57}
& \textbf{0.89}
& 0.07 & \underline{0.13} & \textbf{0.27}
& \textbf{0.16} \\

\bottomrule
\end{tabular*}
\endgroup
\caption{Full end-to-end motion planning comparison on nuScenes.
``nuS.\ FT'' indicates whether a method is fine-tuned on nuScenes.
Bold values indicate the best results, and underlined values indicate
the second-best results.}
\label{tab:nuscenes_full}
\end{table*}

\section{Additional Ablations}
\label{app:ablation}

Unless varied, all ablations use two denoising steps, twelve action tokens, and
six-view (C\(\times\)6) input on NAVSIMv2 navtest.

\paragraph{Number of Action Tokens.}
Table~\ref{tab:inference_config_ablation_app} evaluates the sensitivity of SV-WAM to the number of
predicted action tokens. Using eight tokens shortens the prediction
horizon and slightly reduces EPDMS, while increasing the number to
sixteen introduces additional long-horizon uncertainty. Performance
remains stable across the tested settings, with twelve action tokens
yielding the highest EPDMS in this sensitivity analysis. We use twelve
action tokens in all main experiments.

\begin{table*}[t]
\centering
\begingroup
\small
\setlength{\tabcolsep}{0pt}
\renewcommand{\arraystretch}{0.90}
\begin{tabular*}{\textwidth}{@{}c@{\extracolsep{\fill}}*{10}{c}@{}}
\toprule
Predicted Action Tokens & NC \(\uparrow\) & DAC \(\uparrow\) & DDC \(\uparrow\) & TLC \(\uparrow\) & EP \(\uparrow\) & TTC \(\uparrow\) & LK \(\uparrow\) & HC \(\uparrow\) & EC \(\uparrow\) & EPDMS \(\uparrow\) \\
\midrule
8  & 98.4 & 98.5 & 99.7 & 99.9 & 86.3 & 98.4 & 97.5 & 98.4 & 88.3 & 90.4 \\
12 & 98.6 & 98.8 & 99.6 & 99.9 & 86.8 & 98.3 & 97.8 & 98.4 & 88.4 & \textbf{91.0} \\
16 & 98.5 & 98.2 & 99.7 & 99.8 & 86.9 & 98.1 & 97.9 & 98.3 & 88.1 & 90.5 \\
\bottomrule
\end{tabular*}
\endgroup
\caption{Action-token ablation on NAVSIMv2 navtest (two denoising steps, C\(\times\)6
input). Metrics are reported as percentages.}
\label{tab:inference_config_ablation_app}
\end{table*}

\paragraph{Training-Side Hyperparameters.}
Table~\ref{tab:training_hparam_ablation} reports sensitivity analyses for the drivable-area
regularization weight and the effective fine-tuning batch size.
EPDMS remains stable for small regularization weights:
both $\lambda_{\mathrm{dac}}=0.01$ and $0.05$ achieve 91.0 EPDMS,
whereas increasing the weight to $0.1$ reduces the score to 90.5.
We use $\lambda_{\mathrm{dac}}=0.01$ in all main experiments.
Performance also varies only marginally across effective fine-tuning
batch sizes of 320, 640, and 1280, with 640 yielding the highest score.
We further observe that planning performance is largely insensitive
to the future-video loss weight within the tested range, and use
$\lambda_{\mathrm{act}}:\lambda_{\mathrm{vid}}=1:1$ throughout.

\begin{table}[t]
\centering

\begingroup
\small
\renewcommand{\arraystretch}{0.90}

\begin{minipage}[t]{0.48\columnwidth}
\centering
\begin{tabular*}{\linewidth}{@{}c@{\extracolsep{\fill}}c@{}}
\toprule
\(\lambda_{\mathrm{dac}}\) & EPDMS \(\uparrow\) \\
\midrule
0.01 & \textbf{91.0} \\
0.05 & \textbf{91.0} \\
0.1 & 90.5 \\
\bottomrule
\end{tabular*}\\[3pt]
(a) Drivable-area weight
\end{minipage}
\hfill
\begin{minipage}[t]{0.48\columnwidth}
\centering
\begin{tabular*}{\linewidth}{@{}c@{\extracolsep{\fill}}c@{}}
\toprule
FT batch & EPDMS \(\uparrow\) \\
\midrule
320 & 90.4 \\
640 & \textbf{91.0} \\
1280 & 90.9 \\
\bottomrule
\end{tabular*}\\[3pt]
(b) Fine-tuning batch size
\end{minipage}

\endgroup

\caption{Additional training hyper-parameter ablations on NAVSIMv2 navtest.
(a) Drivable-area regularizer weight \(\lambda_{\mathrm{dac}}\); (b) effective
fine-tuning batch size. Defaults are \(\lambda_{\mathrm{dac}}=0.01\) and batch size
640. EPDMS is reported as a percentage.}
\label{tab:training_hparam_ablation}
\end{table}

\section{Inference-Efficiency Details}
\label{app:latency}

The main paper compares the online latency of
\method with world-model-based baselines. Latency is measured from preprocessed model inputs to the final
trajectory, covering visual encoding/tokenization, the core
generator, and Other, which includes online state and prompt
preparation, tensor transfers, latent/token packing, scheduler
updates, and action decoding or post-processing. For reference, on a single H20 GPU, \method's deployed action-only path is
dominated by six-view VAE encoding (\(\sim\)175\,ms) and two-step denoising
(\(\sim\)151\,ms); enabling the future-video branch raises the trajectory-only latency to
\(848.0\pm1.8\)\,ms, while optional future-video VAE decoding, required only when
future-video outputs are requested, adds a further \(1394.3\pm3.3\)\,ms. We additionally
provide a deployment-oriented breakdown on a single H800 GPU below.

\paragraph{Deployment at 2\,Hz on H800.}
On a single H800 GPU, \method runs the full preprocessed-input-to-trajectory pipeline at roughly
\(2\,\mathrm{Hz}\) in the deployed action-only setting, i.e.\ within the 0.5\,s budget
of a 2\,Hz planner. Table~\ref{tab:speed_h800} breaks down the H800 cost of the
action-only path against the joint action--video path.

\begin{table}[t]
\centering
\begingroup
\small
\setlength{\tabcolsep}{0pt}
\renewcommand{\arraystretch}{0.90}
\begin{tabular*}{\columnwidth}{@{}l@{\extracolsep{\fill}}ccccc@{}}
\toprule
Inference path & \shortstack{Input\\Enc.} & \shortstack{Core\\Gen.} & \shortstack{VAE\\Dec.} & Total & Speedup \\
\midrule
\shortstack[l]{Action-only\\(deployed)} & \(\sim\)53\,ms & \(\sim\)123\,ms & -- & \(\sim\)176\,ms & \(\sim\)4.4\(\times\) \\
\shortstack[l]{Joint action--\\video} & \(\sim\)135\,ms & \(\sim\)201\,ms & \(\sim\)428\,ms & \(\sim\)776\,ms & \(\sim\)1.0\(\times\) \\
\bottomrule
\end{tabular*}
\endgroup
\caption{Inference-speed breakdown on a single H800 GPU (batch size 1, 12 action
tokens, 2-step inference). Total is measured from preprocessed inputs to the outputs
produced by each inference path. Action-only is the deployed path; joint action--video
additionally denoises and decodes future video (the reported total also includes
\(\sim\)12\,ms of other online operations, such as tensor packing).}
\label{tab:speed_h800}
\end{table}

\section{Additional Qualitative Results}
\label{app:additional_qualitative}

\begin{figure*}[t]
\centering
\includegraphics[width=\textwidth]{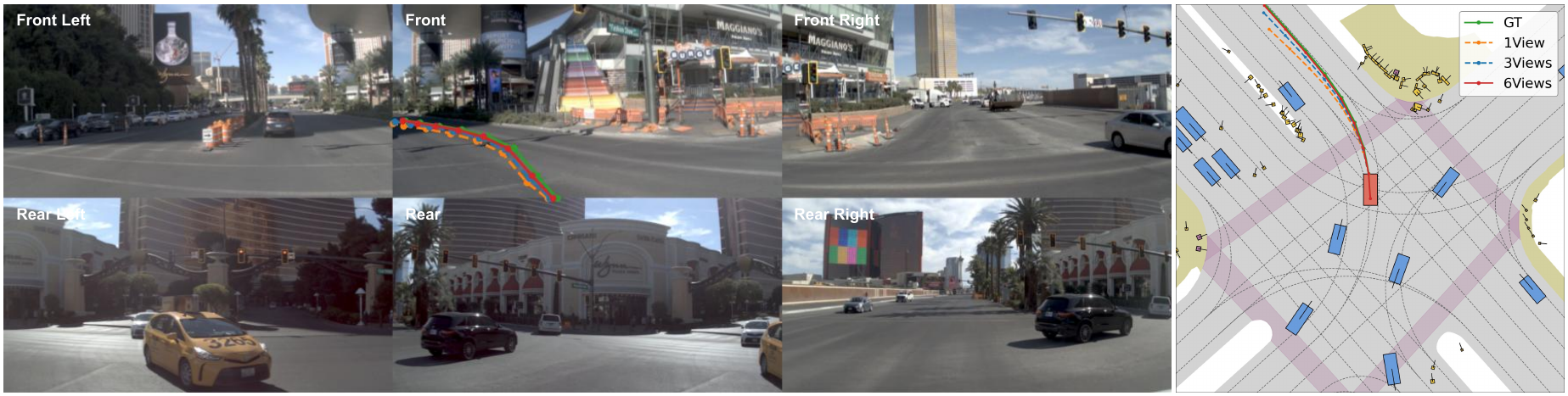}
\caption{Qualitative comparison of camera coverage on NAVSIMv2.
Left: synchronized six-camera observations with ground-truth and predicted
trajectories projected onto the front view. Right: the corresponding BEV
trajectories. Full C\(\times\)6 coverage provides rear and lateral context
unavailable to the reduced-view settings, keeping the predicted trajectory
closer to the ground truth and lane center.}
\label{fig:view_coverage_qualitative}
\end{figure*}

\subsection{Effect of Camera Coverage}

Figure~\ref{fig:view_coverage_qualitative} complements the attention analysis
and the controlled camera-coverage ablation reported in the main paper.
Compared with C\(\times\)1 and C\(\times\)3, full C\(\times\)6 coverage
provides rear and lateral context that captures nearby vehicles and road
boundaries not fully visible under reduced camera coverage. Consequently,
the C\(\times\)6 prediction remains closer to the ground-truth trajectory
and lane center, whereas the reduced-view variants exhibit larger lateral
deviations.

\paragraph{Surround-View Attention Analysis.}
\label{app:attn}

\begin{figure}[t]
\centering
\includegraphics[width=\columnwidth]{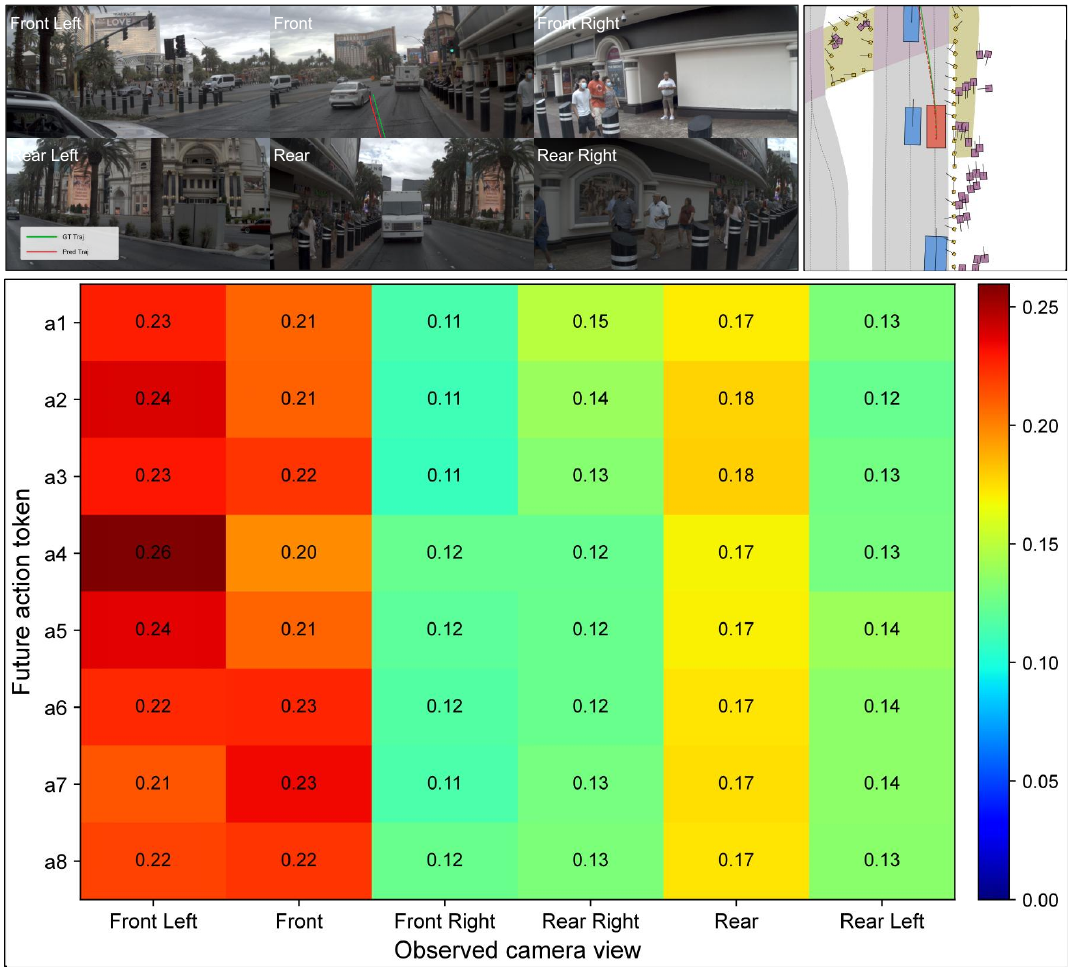}
\caption{Action-to-view attention in a left lane-change, averaged over the 30
transformer blocks. The future action tokens attend most strongly to the front-left and
rear views, where the neighboring vehicle that constrains the maneuver appears,
evidencing that the planner exploits surround-view input.}
\label{fig:attn_views}
\end{figure}

To further examine how \method uses the additional camera views, we visualize the
attention from the future action tokens to the six historical camera views during a
left lane-change maneuver, averaged over the 30 transformer blocks. As shown in
Figure~\ref{fig:attn_views}, instead of relying on the front view alone, the action
tokens place their strongest weight on the front-left and rear views, which contain the
neighboring vehicle that constrains the maneuver. This provides direct evidence that the
action generator reads side and rear context when making lateral decisions, consistent
with the surround-view coverage ablation.

\subsection{Zero-shot In-house Transfer}
\begin{figure}[t]
\centering
\includegraphics[width=\columnwidth]{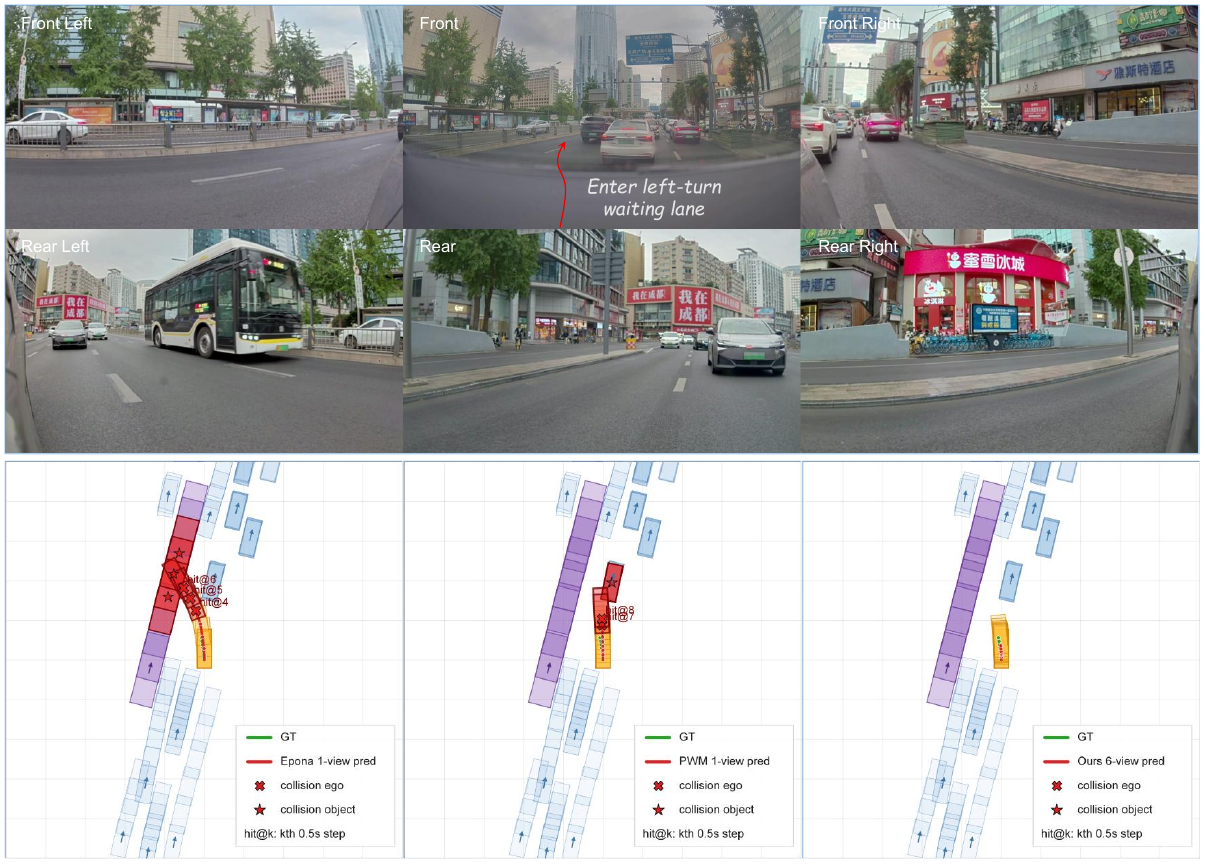}
\caption{Zero-shot qualitative comparison on in-house data when entering a left-turn
waiting lane. Epona (front view only) collides with the rear-left vehicle; PWM avoids
the rear-left vehicle but its zero-shot trajectory drifts substantially from the ground
truth; \method remains closest to the ground-truth trajectory.}
\label{fig:zeroshot_compare}
\end{figure}

Figure~\ref{fig:zeroshot_compare} compares zero-shot planning on an in-house
left-turn waiting-lane scenario. Epona and PWM use their official code and released
checkpoints (PWM uses the NAVSIM-finetuned checkpoint), and none of the methods is
fine-tuned on the target domain. Front-only Epona misses the vehicle approaching from
the rear-left and collides. PWM avoids colliding with the rear-left vehicle, but its
zero-shot trajectory is poor and deviates substantially from the ground truth. In
contrast, \method uses the surround-view observation and remains closest to the ground
truth while clearing the rear-left vehicle.

\subsection{Scenario-wise Planning Examples}
\begin{figure*}[p]
\centering
\includegraphics[height=0.85\textheight,keepaspectratio]{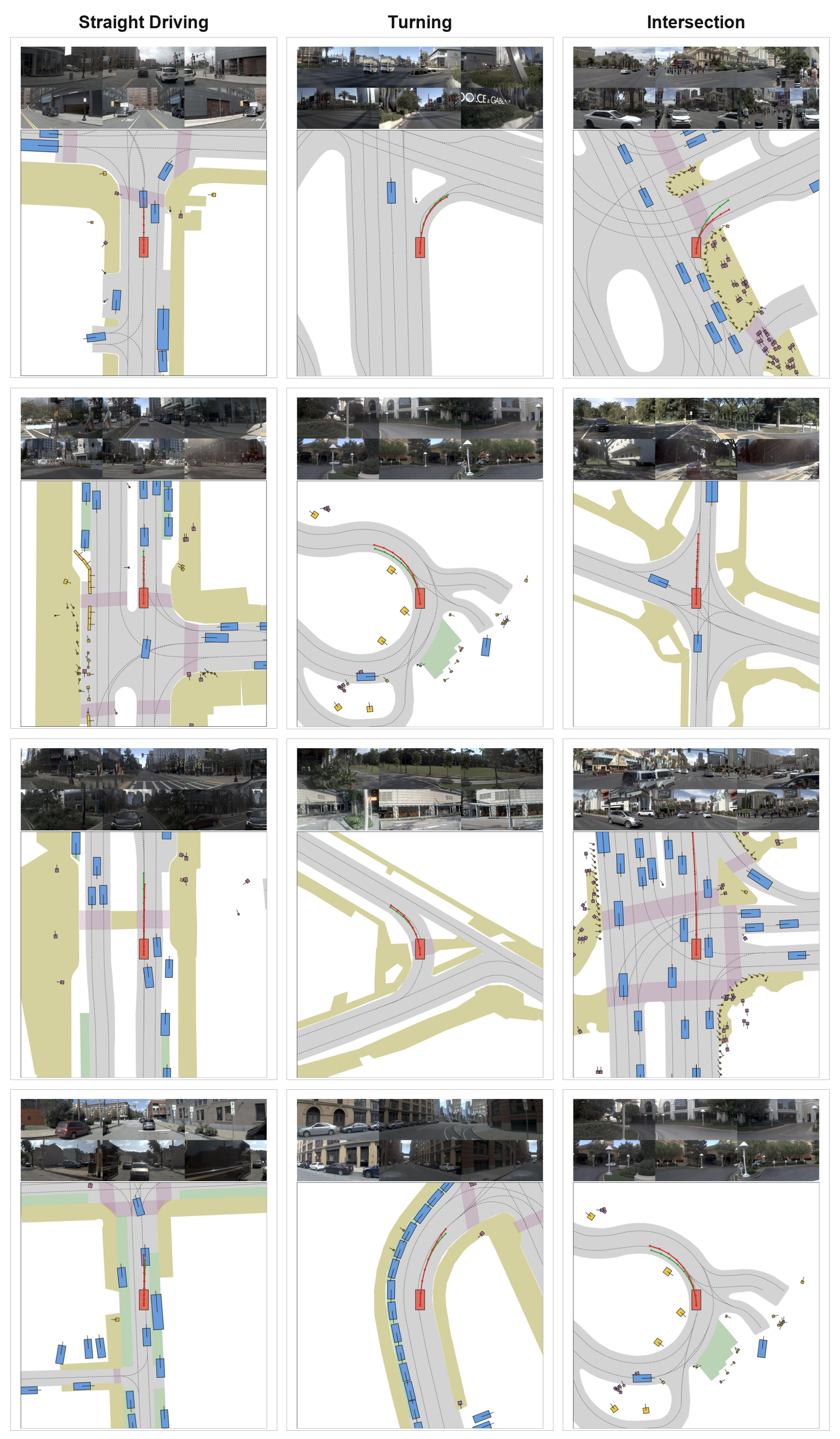}
\caption{Representative high-scoring planning examples grouped by maneuver type.
Columns are straight driving, turning, and intersection scenarios; each column shows
four examples. Green curves denote the ground-truth trajectories, and red curves denote the predicted trajectories.}
\label{fig:qual_success_by_scenario_a}
\end{figure*}

\begin{figure*}[p]
\centering
\includegraphics[height=0.85\textheight,keepaspectratio]{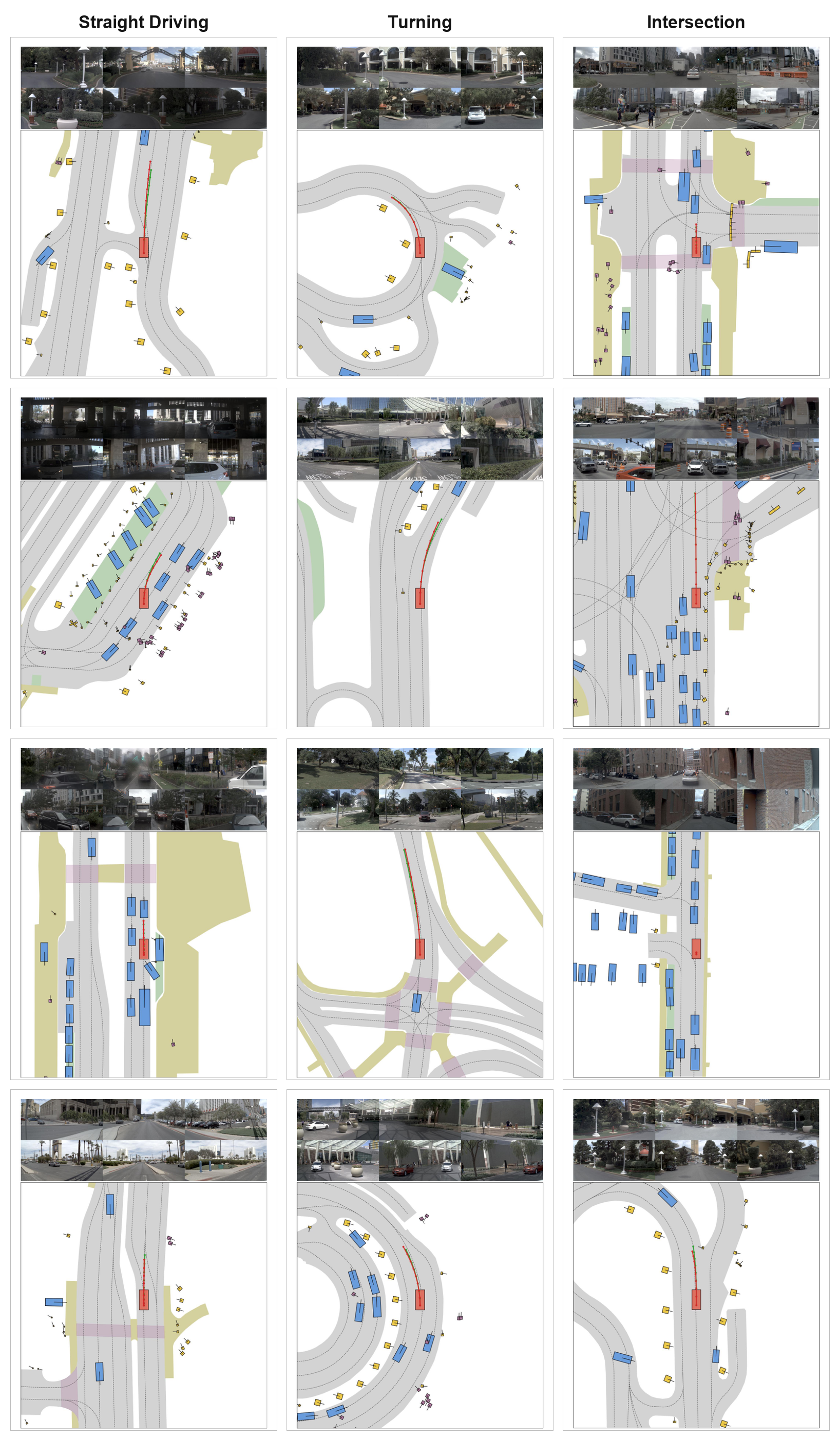}
\caption{Additional high-scoring planning examples grouped by maneuver type, four per
column.}
\label{fig:qual_success_by_scenario_b}
\end{figure*}

Figures~\ref{fig:qual_success_by_scenario_a} and
\ref{fig:qual_success_by_scenario_b} present representative high-scoring trajectories
on the NAVSIMv2 navtest split, across straight driving, turning, and intersection
scenarios. The same action-only
planner remains stable across these common closed-loop maneuvers despite their
different interaction and geometry patterns.

\subsection{Comparison with Other Methods}
\begin{figure*}[!htbp]
\centering
\includegraphics[height=0.88\textheight,keepaspectratio]{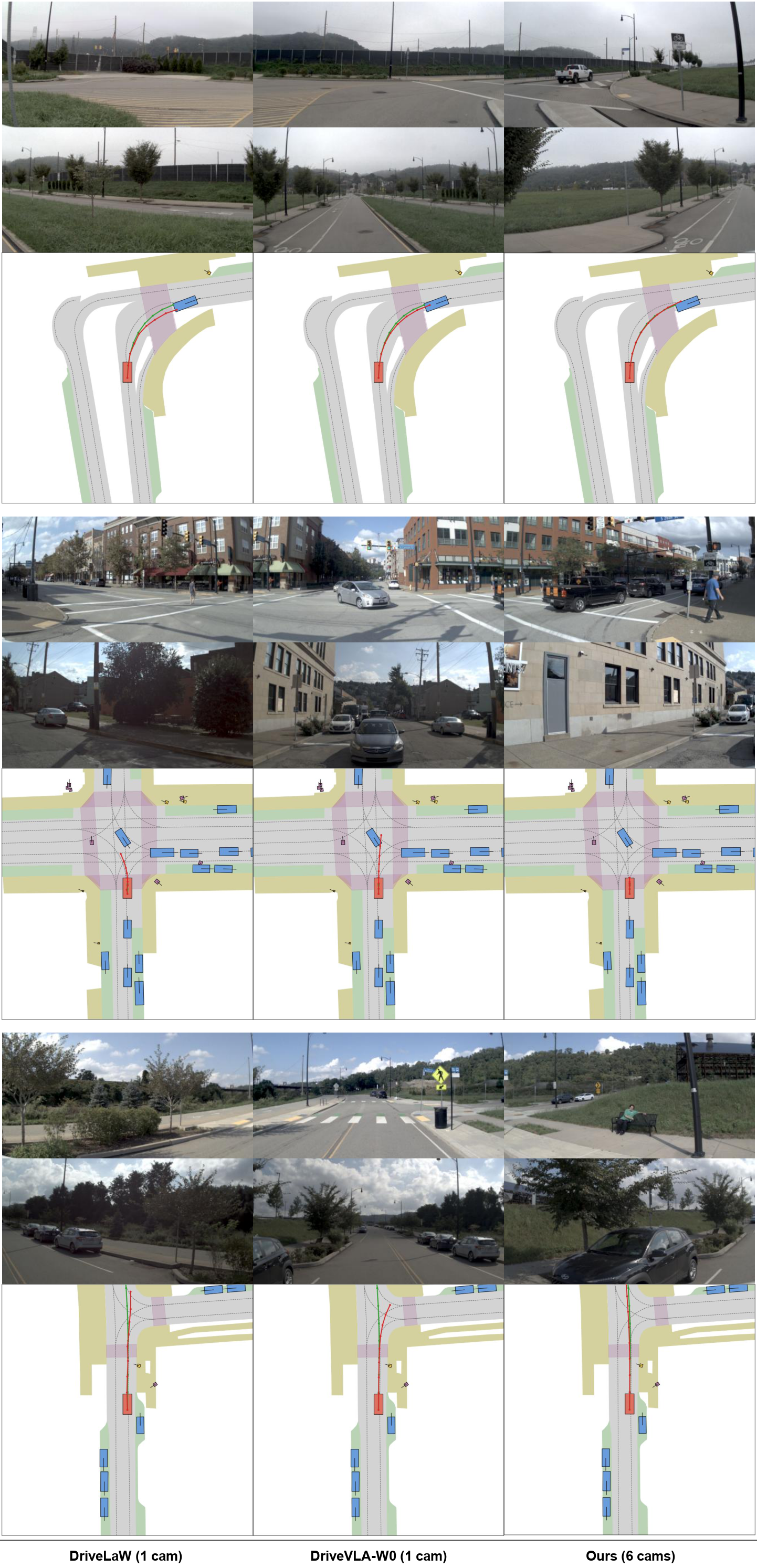}
\caption{Qualitative comparison with front-view baselines. Each row is one scene; the
three columns, labeled at the bottom of the figure, are DriveLaW and DriveVLA-W0
(front camera only, C\(\times\)1) and \method (six-camera surround-view, C\(\times\)6).}
\label{fig:qual_compare_01_03}
\end{figure*}

\begin{figure*}[p]
\centering
\includegraphics[height=0.88\textheight,keepaspectratio]{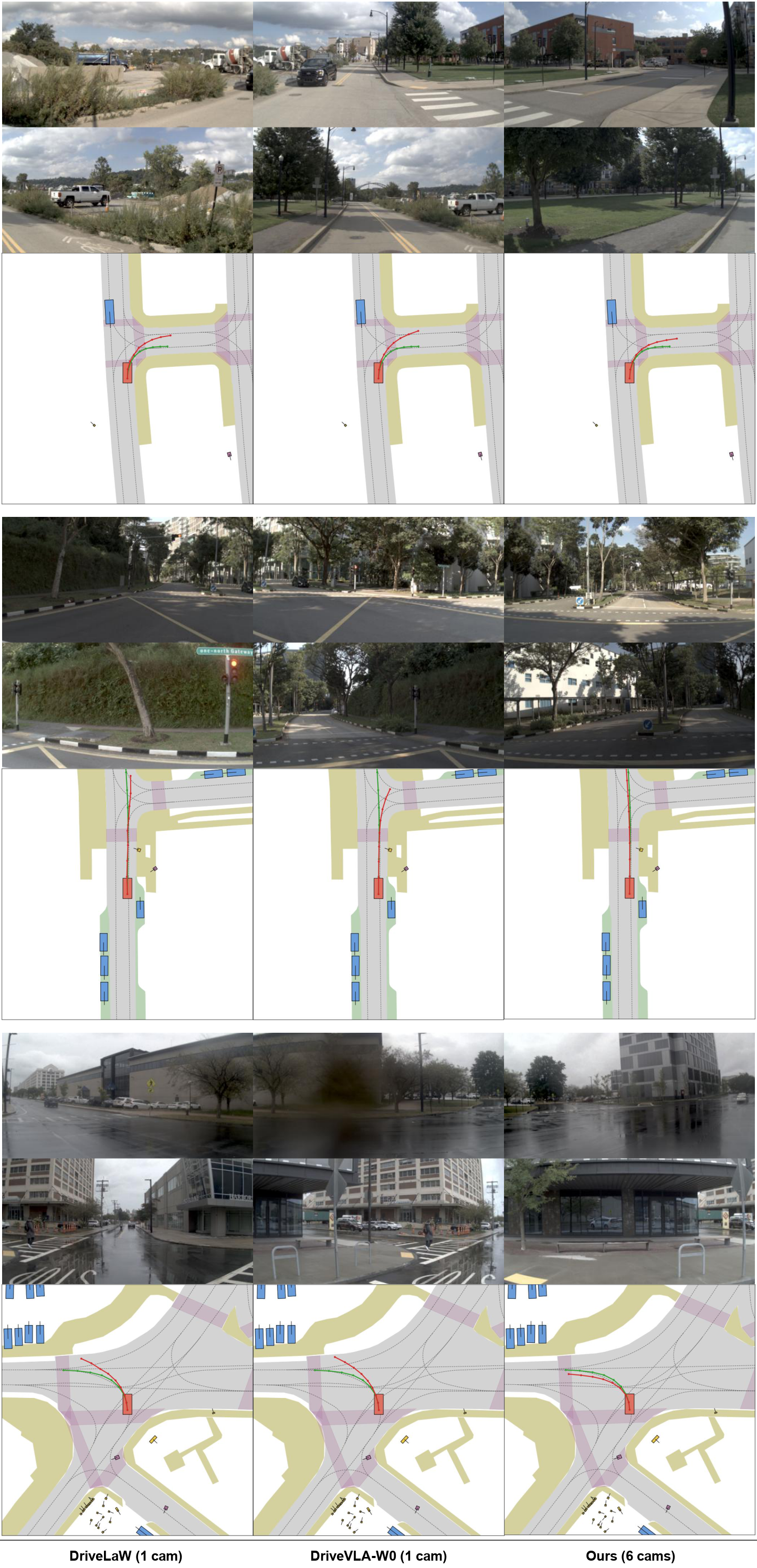}
\caption{Additional qualitative comparison with front-view baselines; same three-column
layout (DriveLaW, DriveVLA-W0, \method) and camera labeling as in
Figure~\ref{fig:qual_compare_01_03}.}
\label{fig:qual_compare_04_06}
\end{figure*}

\begin{figure*}[p]
\centering
\includegraphics[height=0.88\textheight,keepaspectratio]{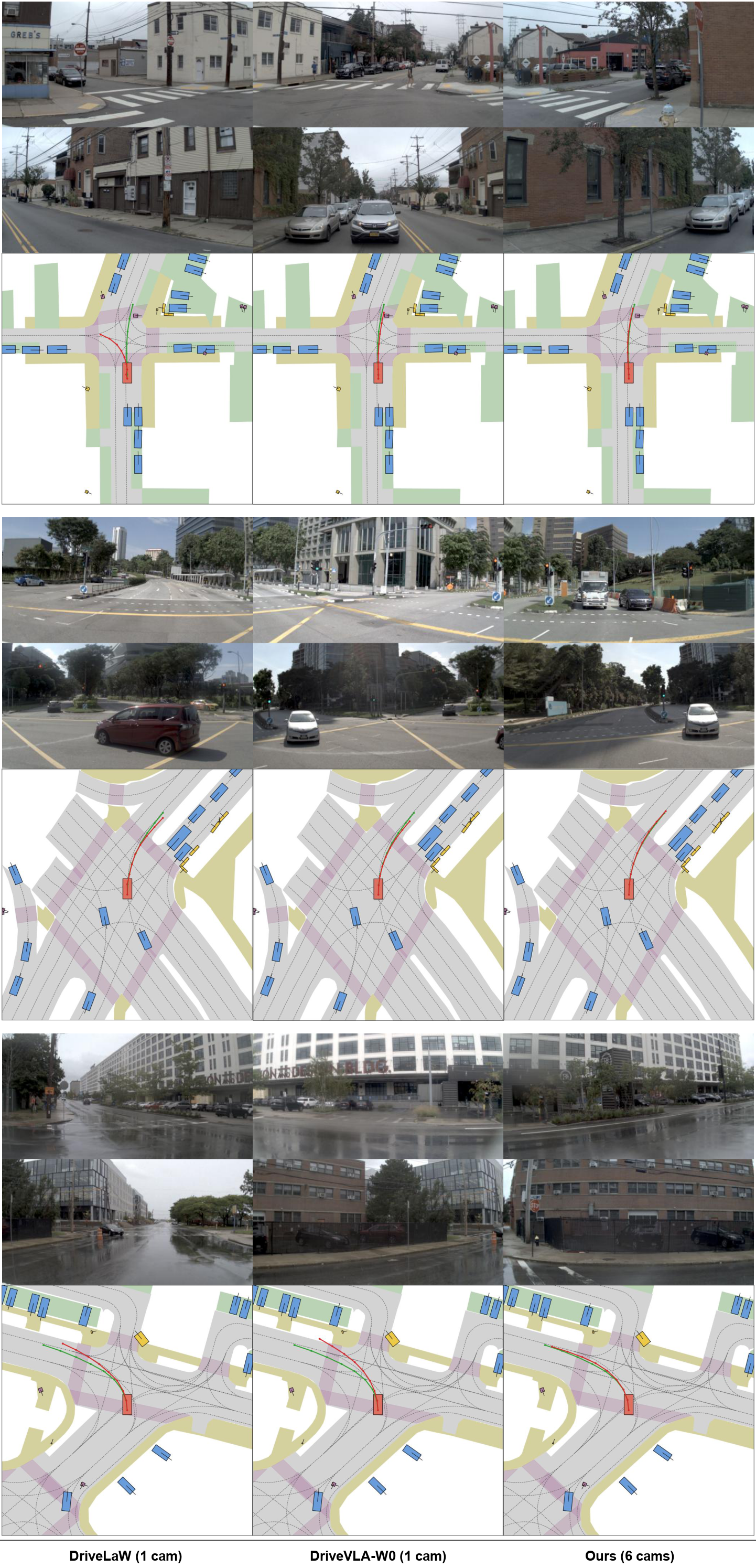}
\caption{Additional qualitative comparison with front-view baselines; same three-column
layout (DriveLaW, DriveVLA-W0, \method) and camera labeling as in
Figure~\ref{fig:qual_compare_01_03}.}
\label{fig:qual_compare_07_09}
\end{figure*}

\begin{figure*}[p]
\centering
\includegraphics[height=0.88\textheight,keepaspectratio]{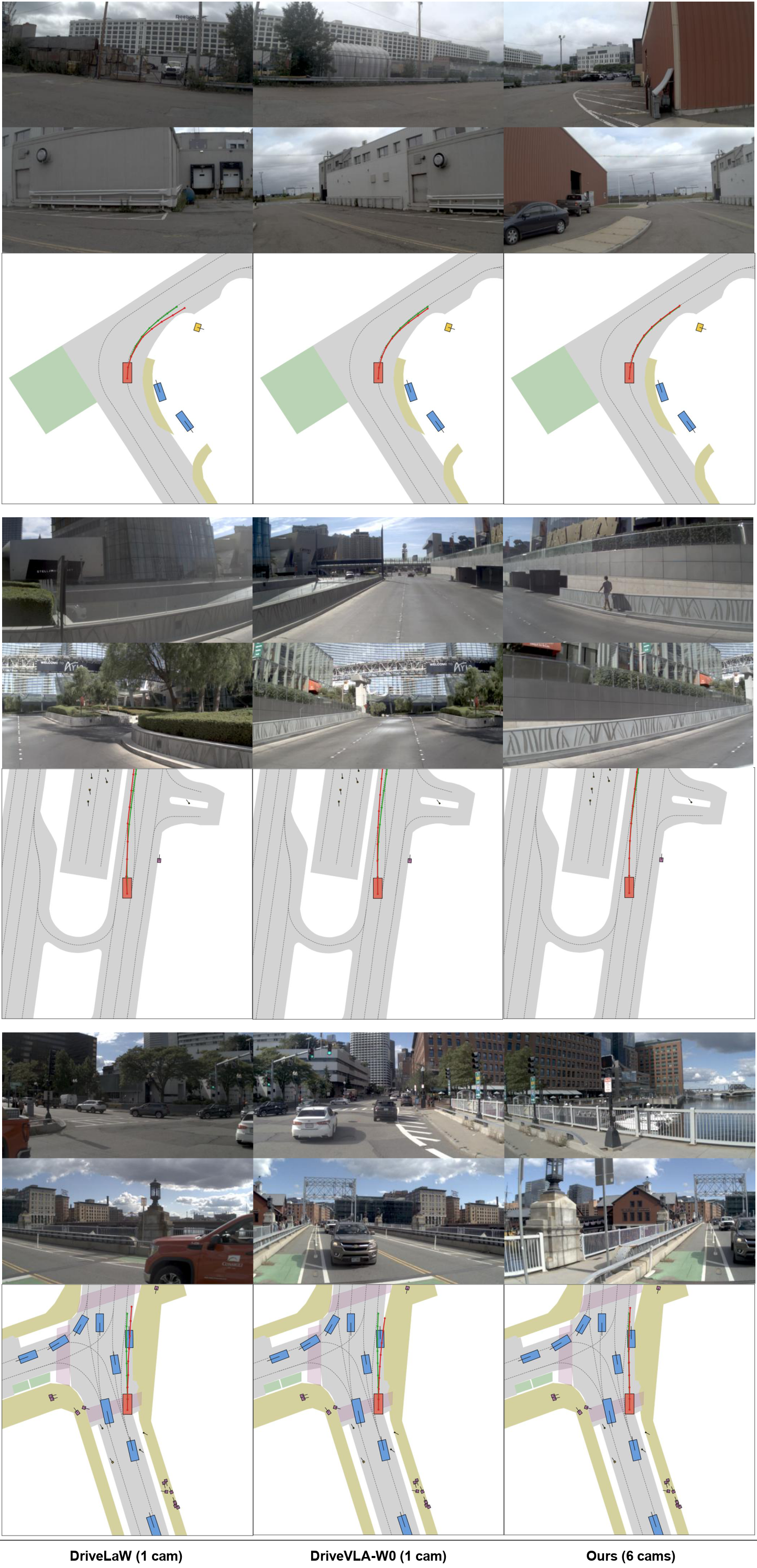}
\caption{Additional qualitative comparison with front-view baselines; same three-column
layout (DriveLaW, DriveVLA-W0, \method) and camera labeling as in
Figure~\ref{fig:qual_compare_01_03}.}
\label{fig:qual_compare_10_12}
\end{figure*}

\begin{figure*}[p]
\centering
\includegraphics[height=0.88\textheight,keepaspectratio]{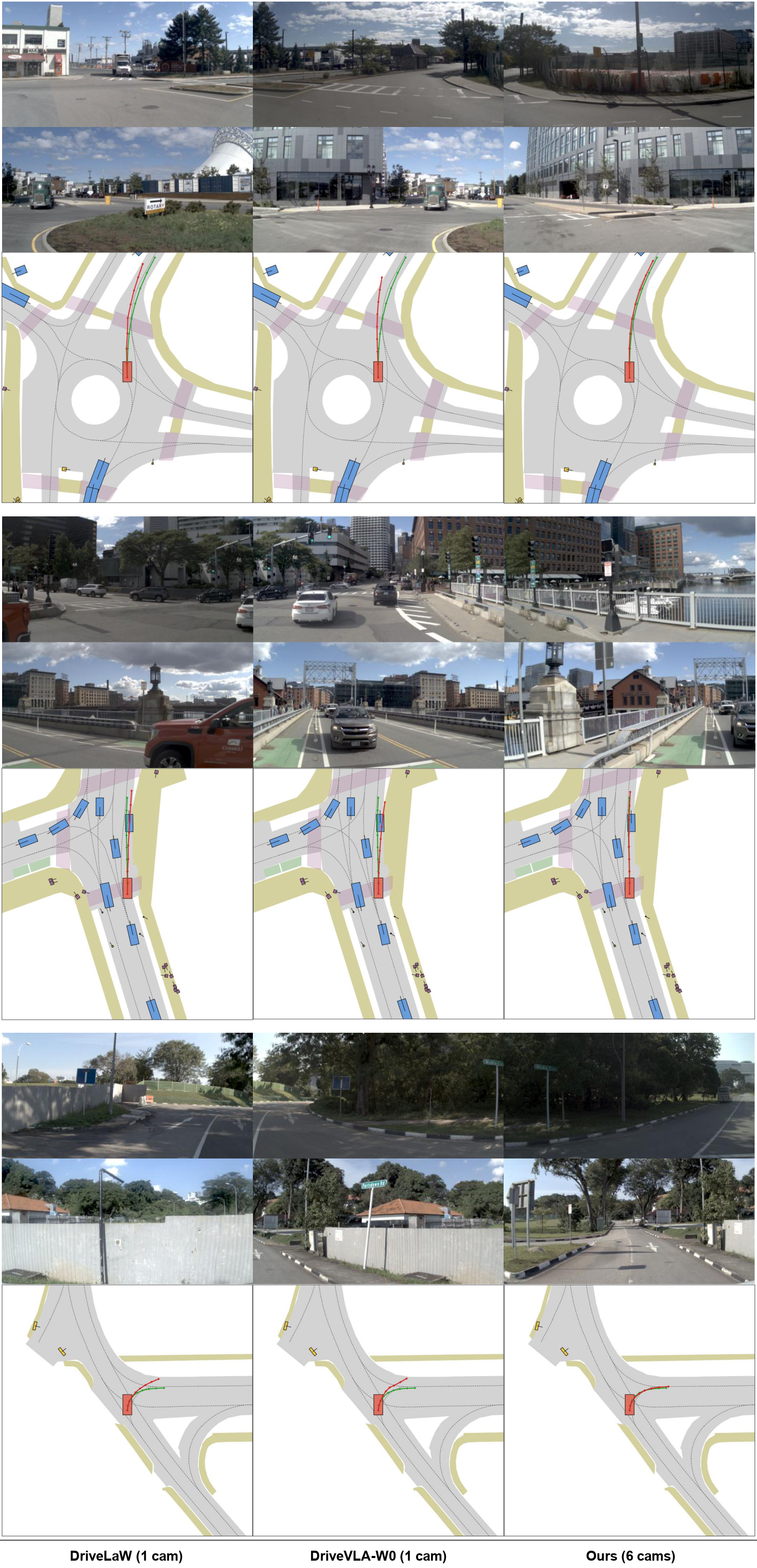}
\caption{Additional qualitative comparison with front-view baselines; same three-column
layout (DriveLaW, DriveVLA-W0, \method) and camera labeling as in
Figure~\ref{fig:qual_compare_01_03}.}
\label{fig:qual_compare_13_15}
\end{figure*}

Figures~\ref{fig:qual_compare_01_03}--\ref{fig:qual_compare_13_15} compare DriveLaW,
DriveVLA-W0, and \method across 15 representative NAVSIMv2 navtest scenarios. Each row is one scene,
and the three columns correspond to DriveLaW, DriveVLA-W0, and \method. DriveLaW and
DriveVLA-W0 use front-view input, whereas \method uses six-camera surround-view input.
The examples include dense intersections, turning maneuvers, and interactions with
nearby traffic.

\subsection{Failure Cases}
\begin{figure*}[t]
\centering
\includegraphics[height=0.88\textheight,keepaspectratio]{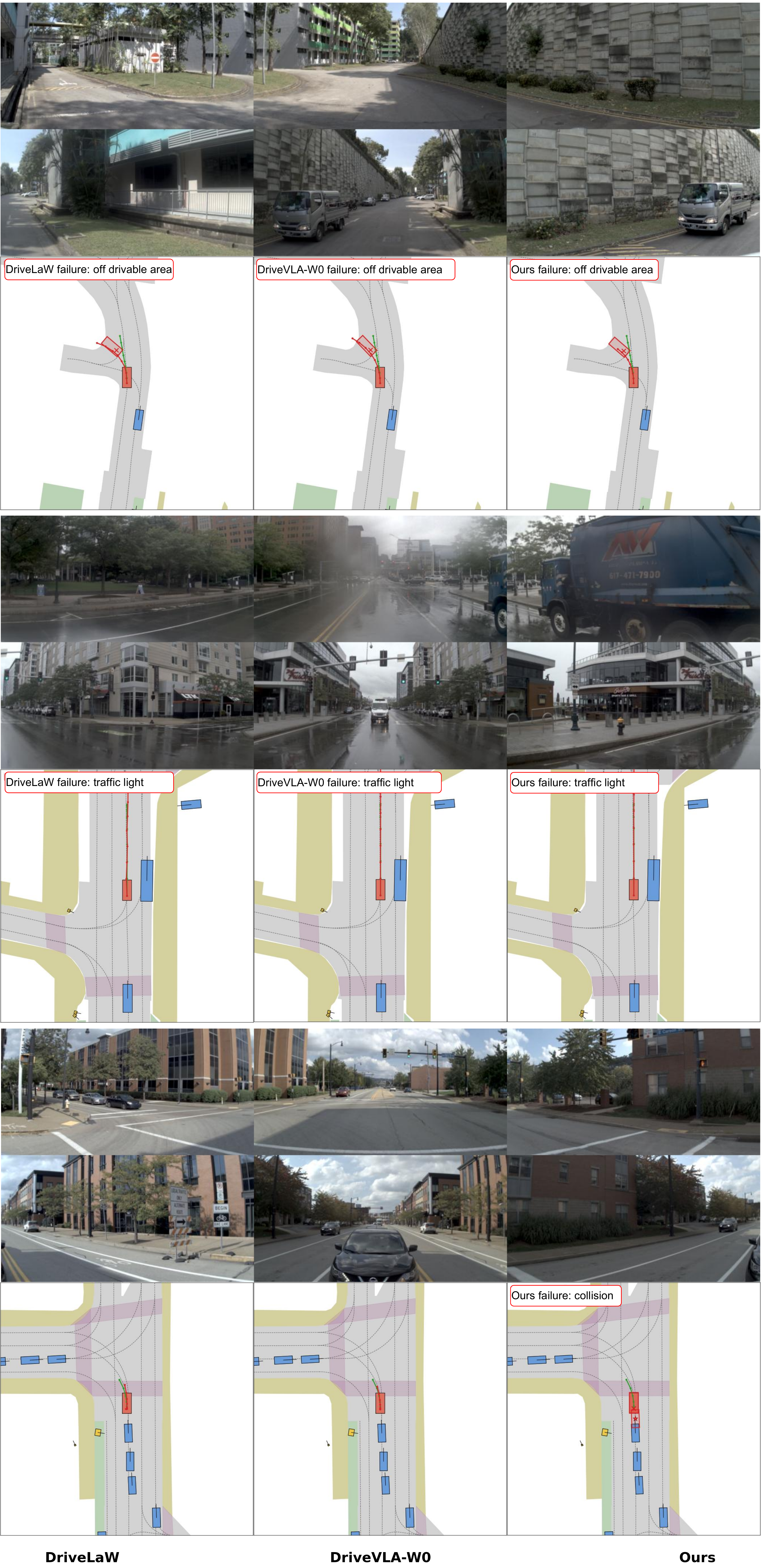}
\caption{Representative failure cases under challenging conditions. Each row compares DriveLaW, DriveVLA-W0, and \method. The three cases correspond to ambiguous left-turn geometry, rain-obscured traffic-signal perception, and overly conservative acceleration after a green signal, respectively.}
\label{fig:failure_cases}
\end{figure*}

Figure~\ref{fig:failure_cases} presents representative failure cases to characterize the remaining limitations of \method. In the first case, the ambiguous two-branch left-turn geometry causes all three methods to select an incorrect path. In the second case, heavy rain obscures the forward view and traffic signal, leading all three methods to fail. In the third case, \method
accelerates too conservatively after the signal turns green and is rear-ended, while DriveLaW and DriveVLA-W0 complete the scenario successfully. These examples suggest that stronger interaction reasoning and more robust uncertainty handling remain important directions for future work.

\section{Limitations and Future Work}
\label{app:limitations}
Although \method achieves strong planning performance with efficient
action-only inference, its approximately 5B-parameter backbone remains
relatively large for deployment on resource-constrained on-board platforms.
We plan to scale training with larger and more diverse
real-world driving data to further improve robustness and generalization,
while exploring knowledge distillation, pruning, and quantization to derive
lightweight variants suitable for edge deployment. We will also extend the
current benchmark-based evaluation to closed-loop testing on real vehicles,
providing a more comprehensive assessment of reliability and practical
deployability under diverse traffic and environmental conditions.

\end{document}